\documentclass[10pt,twocolumn,letterpaper]{article}

\usepackage{cvpr}              
\definecolor{cvprblue}{rgb}{0.21,0.49,0.74}
\usepackage[pagebackref,breaklinks,colorlinks,allcolors=cvprblue]{hyperref}

\usepackage{graphicx}
\usepackage{multirow}
\usepackage{amsthm,amsmath,amssymb}
\usepackage{mathrsfs}
\usepackage{makecell}
\usepackage{bm}
\usepackage{threeparttable}

\def\paperID{*****} 
\def\confName{CVPR}
\def\confYear{2026}

\title{Ultra-Low Bitrate Perceptual Image Compression with Shallow Encoder}

\author{Tianyu Zhang$^{1,2}$ \quad Dong Liu$^{1}$ \quad Chang Wen Chen$^{2}$\\
$^{1}$ University of Science and Technology of China \quad $^{2}$ The Hong Kong Polytechnic University\\
{\tt\small zhangtianyu@mail.ustc.edu.cn, dongeliu@ustc.edu.cn, changwen.chen@polyu.edu.hk}}

\begin{document}
\maketitle

\begingroup
\renewcommand\thefootnote{}
\footnote{This work is supported by the Natural Science Foundation of China under Grant U25B2010, and Hong Kong Research Grants Council (GRF-15229423). (Corresponding author: Dong Liu.)}
\addtocounter{footnote}{-1}
\endgroup

\begin{abstract}
Ultra-low bitrate image compression (below 0.05 bits per pixel) is increasingly critical for bandwidth-constrained and computation-limited encoding scenarios such as edge devices. Existing frameworks typically rely on large pretrained encoders (e.g., VAEs or tokenizer-based models) and perform transform coding within their generative latent space. While these approaches achieve impressive perceptual fidelity, their reliance on heavy encoder networks makes them unsuitable for deployment on weak sender devices. In this work, we explore the feasibility of applying shallow encoders for ultra-low bitrate compression and propose a novel \textbf{A}symmetric \textbf{E}xtreme \textbf{I}mage \textbf{C}ompression (\textbf{AEIC}) framework that pursues simultaneously encoding simplicity and decoding quality. Specifically, AEIC employs moderate or even shallow encoder networks, while leveraging an one-step diffusion decoder to maintain high-fidelity and high-realism reconstructions under extreme bitrates. To further enhance the efficiency of shallow encoders, we design a dual-side feature distillation scheme that transfers knowledge from AEIC with moderate encoders to its shallow encoder variants. Experiments show that AEIC not only outperforms existing methods on rate-distortion-perception performance at ultra-low bitrates, but also delivers exceptional encoding efficiency for 35.8 FPS on 1080P images, while maintaining competitive decoding speed compared to existing methods. Code is available at \href{https://github.com/LuizScarlet/AEIC}{https://github.com/LuizScarlet/AEIC}.
\end{abstract}    
\section{Introduction}
\label{sec:intro}
Image compression is the core of visual communication systems that enable efficient storage and transmission of visual data across various platforms. Traditional compression standards such as JPEG \cite{wallace1991jpeg} and H.266/VVC \cite{bross2021overview} are built upon hand-crafted transforms and quantization, while recent neural image compression \cite{he2022elic, balle2017end, balle2018variational, cheng2020learned, guo2021causal, jiang2023mlicpp} has revolutionized this domain by jointly learning non-linear transforms and latent representations. These methods display superior adaptability and visual quality compared to manually designed standards, especially at moderate bitrates.

However, when operating at ultra-low bitrates where only a few bits are available, image compression faces unique and severe challenges. Such conditions are particularly relevant for bandwidth-constrained and computation-limited senders, including edge devices and IoT terminals. In these cases, the encoder must operate under stringent computation and bitrate budgets. Traditional standards \cite{wallace1991jpeg, sullivan2012overview, bross2021overview} tend to produce severe blurring and blocking artifacts due to the inherent rate-distortion tradeoff. Moreover, their hand-crafted nature prevents them from optimizing directly toward perception, which is critical in human-aligned visual quality assessment at extremely low bitrates.

\begin{figure}[!t]
    \centering
    \includegraphics[width=0.48\textwidth]{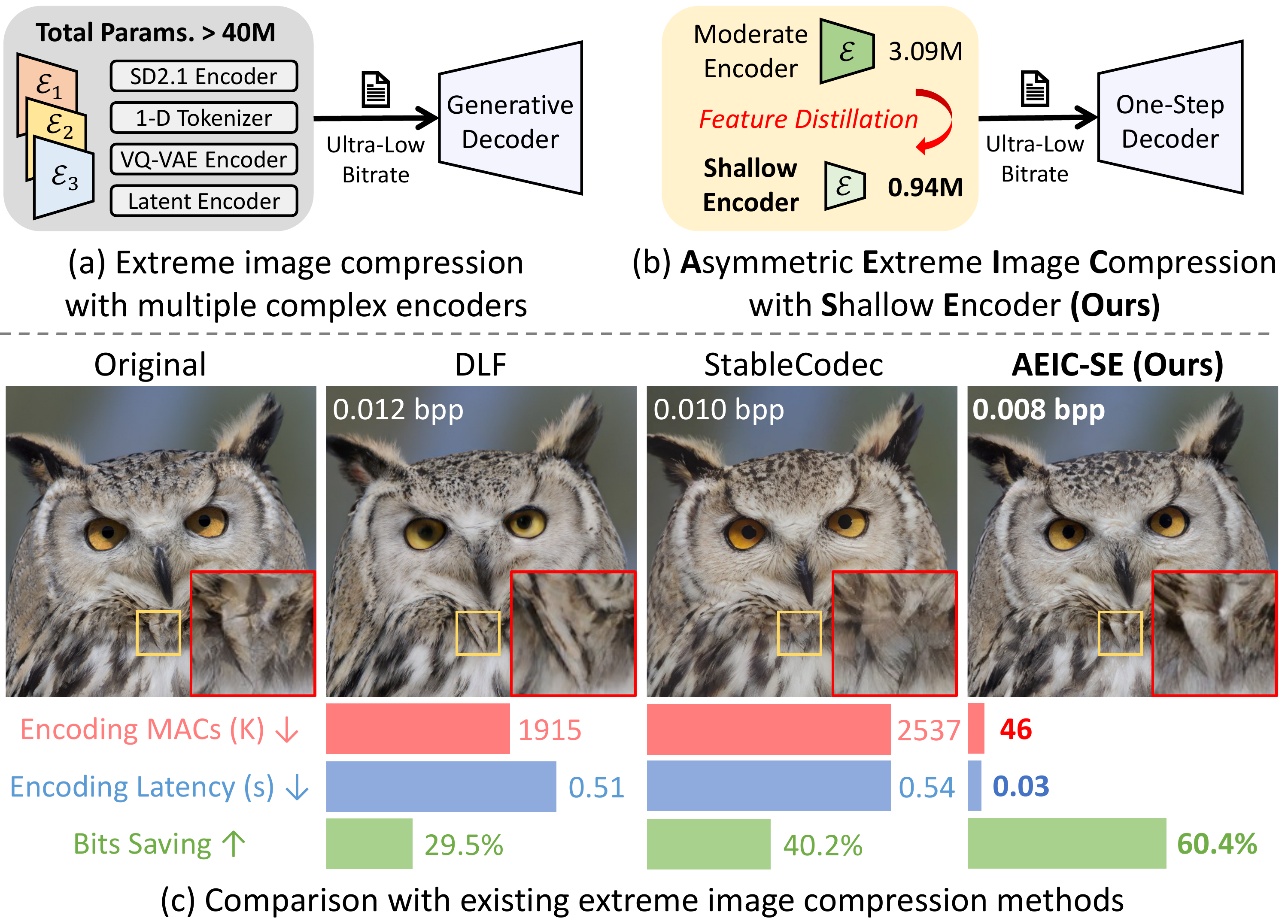}  
    \captionsetup{skip=2pt}
    \caption{(a) Existing extreme image compression methods rely on latent-space transform coding with complex encoders, making them unsuitable for source-limited senders. (b) We propose AEIC, enabling shallow encoders for extreme compression. (c) Our shallow encoder variant AEIC-SE obtains superior performance and encoding complexity compared to advanced methods DLF \cite{Xue_2025_ICCV} and StableCodec \cite{Zhang_2025_ICCV}. Bits saving is calculated by DISTS \cite{ding2020image} on CLIC 2020 Test  \cite{toderici2020clic} compared to GLC \cite{jia2024generative}. We evaluate encoding complexity by MACs per pixel and latency on 1080P images.}
    \label{fig1}
\end{figure}

To address this, modern learning-based perceptual compression frameworks \cite{Zhang_2025_ICCV, Xue_2025_ICCV, jia2024generative, li2024towards, li2025rdeic, korber2024perco} have shifted from pixel-domain reconstruction to latent generative representations. Recent works employ pretrained tokenizers \cite{Zhang_2025_ICCV, jia2024generative} or variational autoencoders (VAEs) \cite{Zhang_2025_ICCV, li2024towards, li2025rdeic, korber2024perco} to map images into a compact latent space, on which transform coding \cite{balle2017end} is applied to achieve extremely low bitrates while maintaining realistic and semantically coherent reconstructions. Generative priors realized via powerful diffusion or transformer decoders further improve perceptual quality under aggressive compression. However, these methods rely on complex pretrained encoders in combination with a secondary latent-space encoder for entropy modeling, resulting in multiple-encoder structures that impose high computational and memory demands. Such architectures hinder deployment on source-limited devices, where encoding speed and model size are critical constraints.

In this work, we explore the feasibility of using shallow encoders for ultra-low bitrate image compression, both theoretically and practically. We analyze the relationship between bitrate and latent variance, showing that as bitrate decreases, the representational complexity inherently diminishes, allowing for simpler encoder designs. Building on this insight, we propose asymmetric extreme image compression (AEIC) overviewed in Fig. \ref{fig1} featuring a moderate or shallow encoder and a generative one-step diffusion decoder. To enhance shallow encoders, we introduce a dual-side feature distillation scheme that transfers both encoder and decoder knowledge from a moderate encoder teacher (AEIC-ME) to its shallow encoder variant (AEIC-SE).

Extensive experiments suggest AEIC achieves state-of-the-art rate-distortion-perception performance under ultra-low bitrates. Our shallow encoder variant AEIC-SE attains the best perceptual metrics (LPIPS \cite{zhang2018unreasonable}, DISTS \cite{ding2020image}, FID \cite{heusel2017gans}, and KID \cite{binkowski2018demystifying}) while maintaining competitive distortion fidelity. Remarkably, it achieves over 35 frames per second (FPS) encoding throughput on 1080P images, with $19\times$ encoding speedup and over $20\%$ bitrate savings compared to previous methods, while sustaining a comparable decoding speed. To the best of our knowledge, this is the first approach to simultaneously achieve state-of-the-art rate-distortion-perception performance and real-time encoding efficiency in ultra-low bitrate settings, paving the way for practical edge-oriented and communication-efficient visual systems. We summarize our contributions as follows:
\begin{itemize}
\item We provide a theoretical and empirical analysis revealing the potential of applying shallow and low-complexity encoders for ultra-low bitrate image compression.
\item We propose asymmetric extreme image compression (AEIC) composed of lightweight encoders and a one-step diffusion decoder. A dual-side feature distillation strategy is introduced to enhance our shallow encoder variant (AEIC-SE) with efficient knowledge transfer.
\item AEIC-SE obtains state-of-the-art perceptual performance at ultra-low bitrates in terms of LPIPS, DISTS, FID, and KID, while preserving strong fidelity. Notably, AEIC-SE attains over 35 FPS for 1080P image encoding and maintains competitive decoding speed.
\end{itemize}
\section{Related Work}

\subsection{Ultra-Low Bitrate Image Compression}

Ultra-low bitrate image compression usually targets extreme ratio below 0.05 bits per pixel (bpp), where bitrate is insufficient to preserve pixel-level fidelity due to the rate-distortion tradeoff. In such cases, compression systems are typically optimized toward human perception \cite{blau2018perception, blau2019rethinking, yan2021perceptual, yan2022optimally} rather than pure distortion. Recent advances in learning-based image compression have enabled substantial perceptual gains in ultra-low bitrates, incorporating generative models such as GAN \cite{agustsson2019generative, agustsson2023multi, muckley2023improving, korber2024egic, agustsson2023multi, he2022po, mentzer2020high}, tokenizers \cite{van2017neural, esser2021taming, yu2024image, jia2024generative, Xue_2025_ICCV}, and more recently diffusion models \cite{li2024towards, xu2024idempotence, yang2024lossy, relic2024lossy, lei2023text+, careil2023towards, hoogeboom2023high, theis2022lossy, Zhang_2025_ICCV, li2025rdeic, sheng2026cadc}. Among these methods, latent-space modeling \cite{Xue_2025_ICCV, jia2024generative, Zhang_2025_ICCV, li2025rdeic, careil2023towards, li2024towards} has emerged as a dominant paradigm, where transform coding is performed not in the pixel domain but within a generative latent space. Leveraging pretrained encoders from models such as Stable Diffusion VAE \cite{Zhang_2025_ICCV, li2025rdeic, careil2023towards, li2024towards} and tokenizers \cite{jia2024generative, Xue_2025_ICCV} provides a structured latent representation that aligns more closely with human perception, facilitating perceptual optimization at extremely low bitrates. However, these methods typically depend on a pretrained generative encoder followed by an additional latent transform encoder, forming a two-stage encoding pipeline. While effective, this design introduces substantial encoding complexity, making such systems impractical for real-world ultra-low bitrate applications where both bandwidth and computational resources are severely constrained at the encoder, including edge devices and IoT terminals.

\begin{figure*}[!ht]
    \centering
    \includegraphics[width=\textwidth]{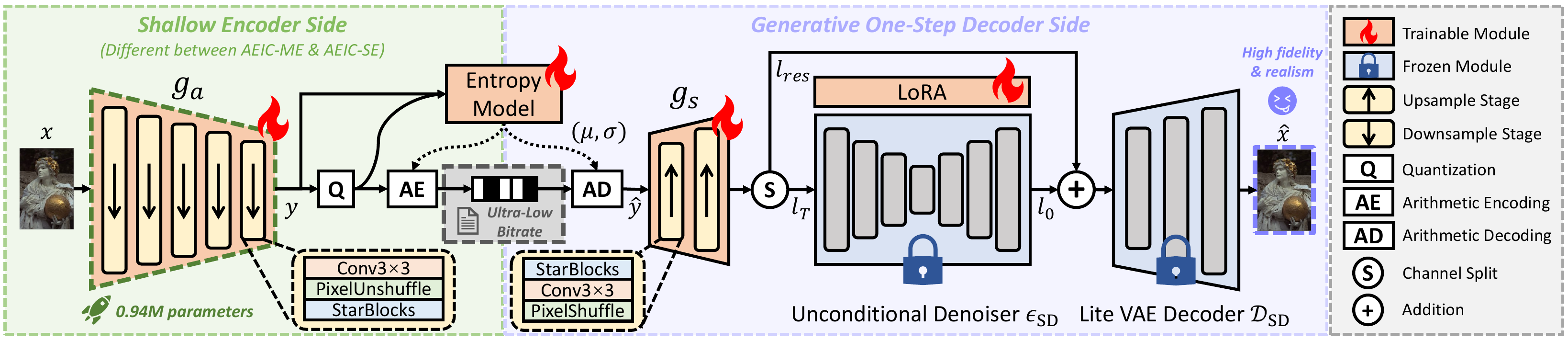}  
    \caption{Pipeline of \textbf{A}symmetric \textbf{E}xtreme \textbf{I}mage \textbf{C}ompression with \textbf{S}hallow \textbf{E}ncoder (\textbf{AEIC-SE}). At the encoder side, we build the analysis transform $g_a$ directly in the pixel space with efficient convolution network (StarNet \cite{ma2024rewrite}). The input image $x$ is first transformed into a compact latent $y$ at a spatial compression ratio of 32, then quantized and entropy coded into $\hat{y}$ by a quadtree-partition \cite{li2023neural} entropy model \cite{minnen2018joint} using ultra-low bitrates. At the decoder side, $\hat{y}$ undergoes the synthesis transform $g_s$, one-step denoiser $\epsilon_{\mathrm{SD}}$ and VAE decoder $\mathcal{D}_{\mathrm{SD}}$ to produce the reconstruction $\hat{x}$. Note that AEIC-ME and AEIC-SE only differ in $g_a$ and the entropy model configuration (Table \ref{config}).}
    \label{framework}
\end{figure*}

\subsection{Real-Time Neural Image Compression}

One major issue in neural coding is the practical complexity. Recent works have introduced architecture \cite{jia2025towards, rippel2021elf, van2024mobilenvc}, network distillation \cite{wang2023evc}, and entropy coding innovations \cite{he2022elic, he2021checkerboard, li2023neural} for practicality. Lightweight decoder designs \cite{yang2023computationally} such as shallow architectures combined with iterative encoding, have also been explored to balance complexity and RD performance. Notably, EVC \cite{wang2023evc} demonstrates that carefully designed architectures with sparsity-based mechanisms can achieve real-time encoding and decoding (e.g., 30 FPS on GPUs) while rivaling traditional codec like VTM \cite{bross2021overview}. Despite these progresses, most real-time techniques are designed primarily for distortion fidelity at moderate bitrates. In contrast, achieving real-time performance at ultra-low bitrates remains largely unexplored due to the inherent computational overhead of generative models.

\section{Method}

\subsection{Asymmetric Extreme Coding Pipeline}
The overall framework of AEIC-SE is illustrated in Fig. \ref{framework}. Given an input image $x$, we directly apply a shallow transform encoder (the analysis transform $g_a$) to compress $x$ into a compact latent representation $y$. The latent $y$ is then quantized into $\hat{y}$, whose Gaussian entropy parameters $(\mu, \sigma)$ for arithmetic coding are estimated by an entropy model consisting of a hyperprior module \cite{balle2018variational} ($h_a$ and $h_s$ as detailed in Fig. \ref{distill}) and a quadtree-partition context model \cite{li2023neural}. At the decoder side, we finetune Stable Diffusion Turbo (SD-Turbo) \cite{sauer2024adversarial} with LoRA \cite{hu2021lora} for one-step decoding. $\hat{y}$ then undergoes the synthesis transform $g_s$, unconditional one-step denoiser $\epsilon_{\mathrm{SD}}$ and a lite VAE decoder $\mathcal{D}_{\mathrm{SD}}$ \cite{chen2025adversarial} to reconstruct $\hat{x}$. The main coding procedure can be formulated as follows (the autoregressive entropy coding for $\hat{y}$ with the entropy model is abbreviated for convenience):
\begin{align}
  \mathrm{Encoding}:\ &y=g_{a}(x) \\
  \mathrm{Quantization}:\ &\hat{y}=\mathrm{quantize}[y-\mu]+\mu \label{eq2}\\
  \mathrm{Decoding}:\ &l_T,l_{res}=\mathrm{split}[g_s(\hat{y})] \label{eq3}\\
  &l_0=\epsilon_{\mathrm{SD}}(l_{T}), \label{eq4}\\
  &\hat{x}=\mathcal{D}_{\mathrm{SD}}(l_{0}+l_{res}) \label{eq5}
\end{align}

\begin{table}[!t]
    \centering
    \setlength{\tabcolsep}{1.0mm}
    \caption{Encoder side model configurations. By adjusting the stage depth (the number of StarBlocks) and dimension in each encoder downsample stage, and those in the entropy model, we build AEIC-ME and AEIC-SE with different analysis transform $g_a$ and entropy model settings, while keeping the same decoder side. Detailed network structures are provided in the supplementary.}
    \label{config}
    \small
    \begin{tabular}{c|c|c|c|c}
        \Xhline{1.0pt}
        \multirow{2}{*}{Model} & \multicolumn{2}{c|}{$g_a$ (each downsample stage)} & \multicolumn{2}{c}{Entropy Model} \\
        \cline{2-5}
        ~ & Depth & Dimension & Depth & Dimension \\
        \hline
        AEIC-ME & 2 & (64, 128, 192, 256, 320) & 4 & 960 \\
        AEIC-SE & 1 & (32, \ \ 64, 128, 192, 256) & 3 & 512 \\
        \Xhline{1.0pt}
    \end{tabular}
\end{table}

\begin{figure}[!t]
    \centering
    \includegraphics[width=0.48\textwidth]{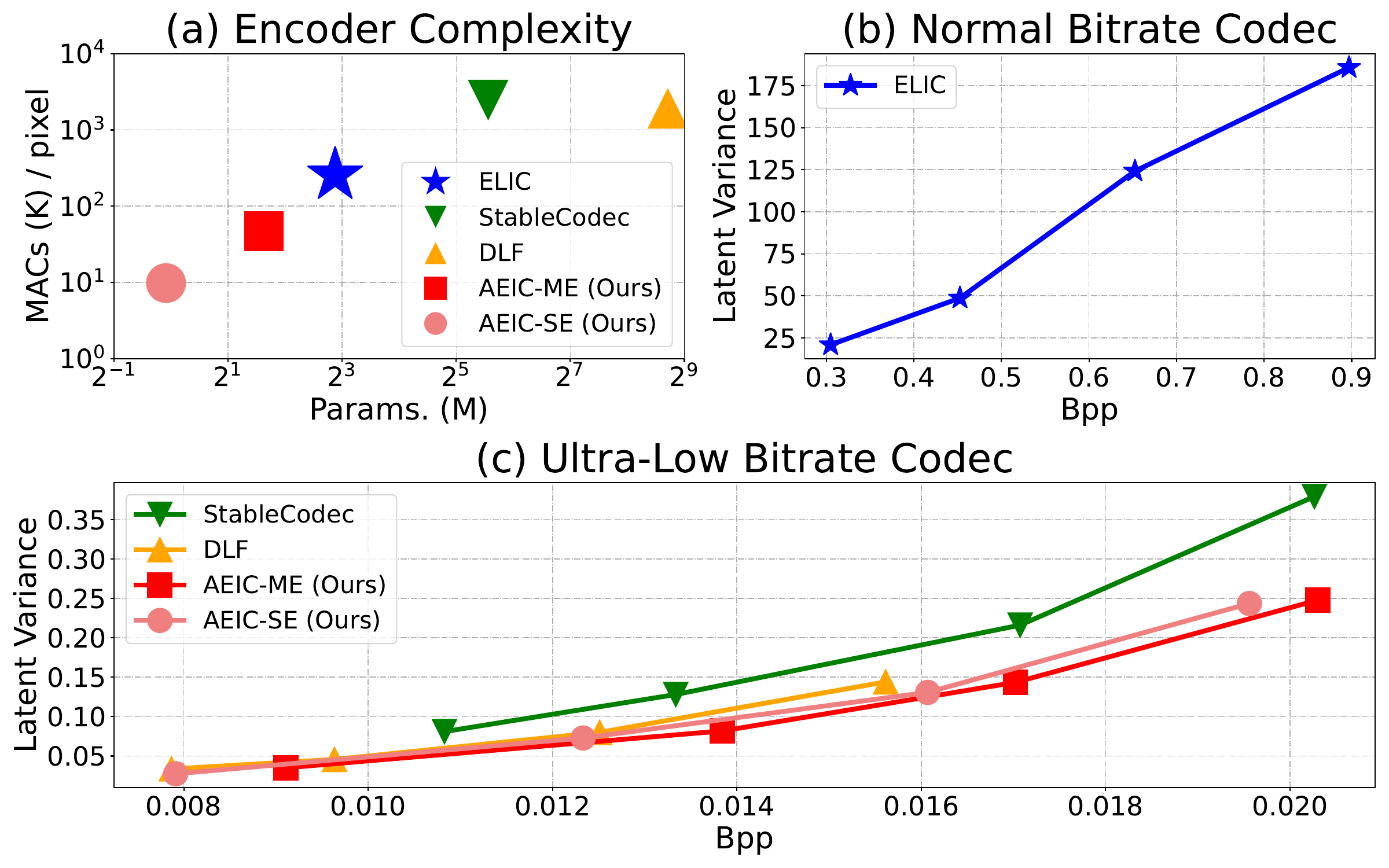}  
    \captionsetup{skip=2pt}
    \caption{Potential of applying shallow encoders for ultra-low bitrate image compression. Latent variance diminishes sharply as bitrate decreases, allowing shallow encoders to express this data range at ultra-low bitrates. (a) Encoder (analysis transform $g_a$) complexity. (b) Latent variance of normal bitrate codec ELIC \cite{he2022elic}. (c) Latent variance of ultra-low bitrate codec StableCodec \cite{Zhang_2025_ICCV}, DLF \cite{Xue_2025_ICCV} and AEIC. Results are averaged on Kodak \cite{kodak}.} 
    \label{encoder}
\end{figure}

\subsubsection{Shallow Transform Encoder for Extreme Bitrate}
Unlike existing methods \cite{Zhang_2025_ICCV, Xue_2025_ICCV, jia2024generative, li2024towards, li2025rdeic, korber2024perco} that rely on complex encoders, we investigate the potential of shallow transform encoder in ultra-low bitrate image compression. 

We begin by analyzing the theoretical relationship between bitrate and encoder complexity in discrete representations. Consider a discrete codebook $C = \{c_1, c_2, \dots, c_M\}$, the maximum representable bitrate $\mathcal{R}$ of such a codebook is $\log_2 M$, which corresponds to the maximum entropy condition. Encoding a signal using this codebook, if using an exhaustive search (without any assumption on the codeword structure), leads to a computational complexity of $\mathcal{O}(M)=\mathcal{O}(2^\mathcal{R})$. Therefore, as the bitrate $\mathcal{R}$ decreases, the required search space and encoding complexity reduce exponentially. This observation suggests that ultra-low bitrate allows for low-complexity encoding process in the discrete representation domain.

We next extend the analysis to the continuous case. For a continuous Gaussian latent variable $z \sim \mathcal{N}(\mu, \sigma^2)$, its differential entropy is defined as $h(z) = \frac{1}{2} \log (2 \pi e \sigma^2)$, showing that the entropy, namely the bitrate, depends solely on the variance $\sigma^2$. Consequently, an ultra-low bitrate codec implies that the latent variance must be small. As shown in Fig. \ref{encoder} (b) and (c), latent variance of neural image codec diminishes sharply as bitrate decreases. A small latent variance in the continuous domain restricts the range of probably sampled values, which is analogous to a discrete codebook with fewer elements. This analogy suggests that ultra-low bitrate compression does not necessarily require a deep or computationally expensive encoder, since the latent information to be encoded is inherently compact.

Based on this insight, we implement two lightweight encoders based on StarNet \cite{ma2024rewrite} with different network configurations in Table \ref{config}, denoted as AEIC-ME (moderate encoder, 3.09M) and AEIC-SE (shallow encoder, 0.94M). Fig. \ref{encoder} (a) indicates the advantages of AEIC on encoder complexity compared to representative methods. In Fig. \ref{encoder} (c), we further suggest that AEIC-ME and AEIC-SE can produce latent representations with similar variance range to existing ultra-low bitrate methods \cite{Zhang_2025_ICCV, Xue_2025_ICCV} using complex encoders.

\subsubsection{Generative Decoder with One-Step Diffusion}

To enhance the perceptual quality of reconstructions under ultra-low bitrates, we construct a generative decoder based on one-step diffusion (Eq. \ref{eq3}-\ref{eq5}), which strikes a balance between reconstruction fidelity and decoding efficiency.

Following StableCodec \cite{Zhang_2025_ICCV}, we employ a dual-branch decoding structure to improve one-step diffusion performance. Specifically, the separated latent representations, $l_T$ for texture generation and $l_{\mathrm{res}}$ for structural residuals, are decoded by a one-step denoiser $\epsilon_{\mathrm{SD}}$ and then fused through element-wise addition before passing into the VAE decoder $\mathcal{D}_{\mathrm{SD}}$. Unlike \cite{Zhang_2025_ICCV}, which utilizes an explicit auxiliary decoder independent of the synthesis transform $g_s$, our approach integrates both decoding branches directly into $g_s$, thereby enhancing cross-branch interaction and decoding efficiency. The two latents, $l_T$ and $l_{\mathrm{res}}$, are split from a shared output latent as formulated in Eq. \ref{eq3}.

In text-to-image synthesis, textual prompt $c$ and timestep $T$ are typically required to condition the diffusion model. For one-step generation \cite{sauer2024adversarial}, the denoising process from a noisy latent $l_T$ to a clean latent $l_0$ can be expressed as:
\begin{equation}
  l_0 = [l_{T} - \sqrt{1-\bar{\alpha}_{T}} \cdot \epsilon_{\mathrm{SD}}(l_{T}, c, T)]/\sqrt{\bar{\alpha}_{T}}
  \label{eq6}
\end{equation}
where $T$ is often set to 999 and $\bar{\alpha}_{T}$ denotes the cumulative DDPM \cite{ho2020denoising} noise schedule $\{\bar{\alpha}_t\}$.  
However, in the image compression setting, transmitting textual prompts introduces additional bitrate overhead. Recent findings \cite{vonderfecht2025lossy} show that such prompts contribute negligible reconstruction improvement compared to the information already contained in the latent codes. In practice, existing methods \cite{Zhang_2025_ICCV, zhang2024degradation} typically fix the text input to generic phrases (e.g., \textit{“a high-resolution, 8K, ultra-realistic image”}) during LoRA fine-tuning, disregarding the prompt’s semantic role.

Motivated by these observations, we remove all prompt and timestep dependencies from the denoiser $\epsilon_{\mathrm{SD}}$ and fine-tune the remaining modules using LoRA in an end-to-end manner to better exploit the generative prior for perceptual reconstruction. Concretely, we prune the original SD-Turbo denoiser \cite{sauer2024adversarial} into an unconditional format by removing text encoders, timestep embeddings, and all cross-attention layers, following the practice of \cite{chen2025adversarial}. As a result, the one-step denoising process in Eq. \ref{eq6} simplifies into the direct decoding formulation described in Eq. \ref{eq4}. Finally, since the VAE decoder $\mathcal{D}_{\mathrm{SD}}$ becomes the computational bottleneck for decoding latency \cite{Zhang_2025_ICCV}, we replace the original one in SD-Turbo with a lightweight variant by pruning 50\% of its channels \cite{chen2025adversarial}, achieving a more efficient decoding process.

\subsection{Knowledge Distillation for Shallow Encoder}
\label{sec:distill}

Due to the limited capacity of the shallow encoder, training AEIC-SE from scratch results in a significant performance drop compared with AEIC-ME, as shown in Table \ref{ablation2}. To mitigate this issue, we propose to transfer knowledge from AEIC-ME (teacher) to AEIC-SE (student), thereby enhancing the shallow encoder’s representational ability for ultra-low bitrate image compression. In this section, we first analyze the training process of AEIC-ME, then introduce the dual-side feature distillation strategy for AEIC-SE.

\begin{figure}[!t]
    \centering
    \includegraphics[width=0.48\textwidth]{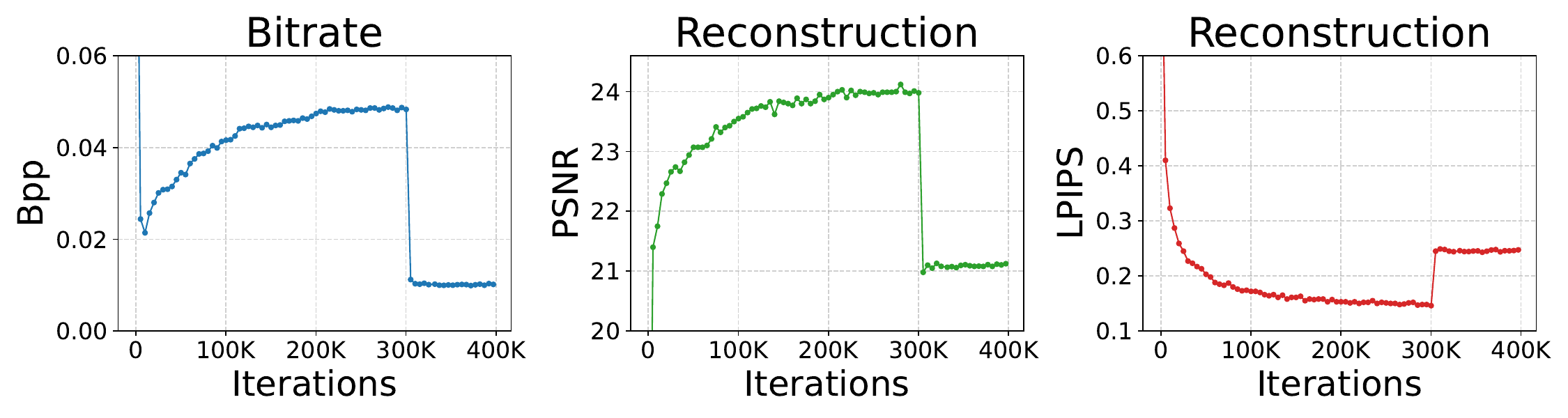}  
    \captionsetup{skip=2pt}
    \caption{Bitrate and reconstruction evolution of AEIC-ME using the two-stage implicit bitrate pruning strategy. The training begins with a relaxed bitrate constraint ($\lambda=1$), then drops to an ultra-low bitrate constraint ($\lambda=16$) at 300K iterations and converges rapidly. Results are averaged on Kodak every 5K iterations.}
    \label{training}
\end{figure}

\noindent\textbf{Analysis on Implicit Bitrate Pruning.} To ensure stable optimization under ultra-low bitrates (below 0.05 bpp), we adopt the two-stage implicit bitrate pruning strategy \cite{Zhang_2025_ICCV}. Specifically, models are first trained with a relaxed bitrate target (0.05 bpp) and subsequently finetuned toward more extreme target bitrates (0.005–0.035 bpp). Fig. \ref{training} visualizes this training process on AEIC-ME in terms of bitrate evolution and reconstruction quality.

In the first stage, since the codec is randomly initialized, it requires a long convergence period. During this phase, the bitrate gradually increases from 0.02 bpp to 0.05 bpp while the reconstruction quality steadily improves. In the second stage, the constraint becomes much more aggressive and the bitrate drops rapidly to the target range, leading to a corresponding decline in the reconstruction performance.  

From the encoder’s perspective, the first stage allows it to explore expressive feature transforms under a relatively loose bitrate constraint. In contrast, the second stage increases bitrate constraints, posing severe degradation to decoding performance. To effectively transfer knowledge from the pretrained AEIC-ME to AEIC-SE, we design two separate distillation objectives $\mathcal{L}_{enc}$ and $\mathcal{L}_{dec}$, which supervise the shallow encoder training in the first stage and the decoder finetuning in the second stage, respectively.

\begin{figure}[!t]
    \centering
    \includegraphics[width=0.48\textwidth]{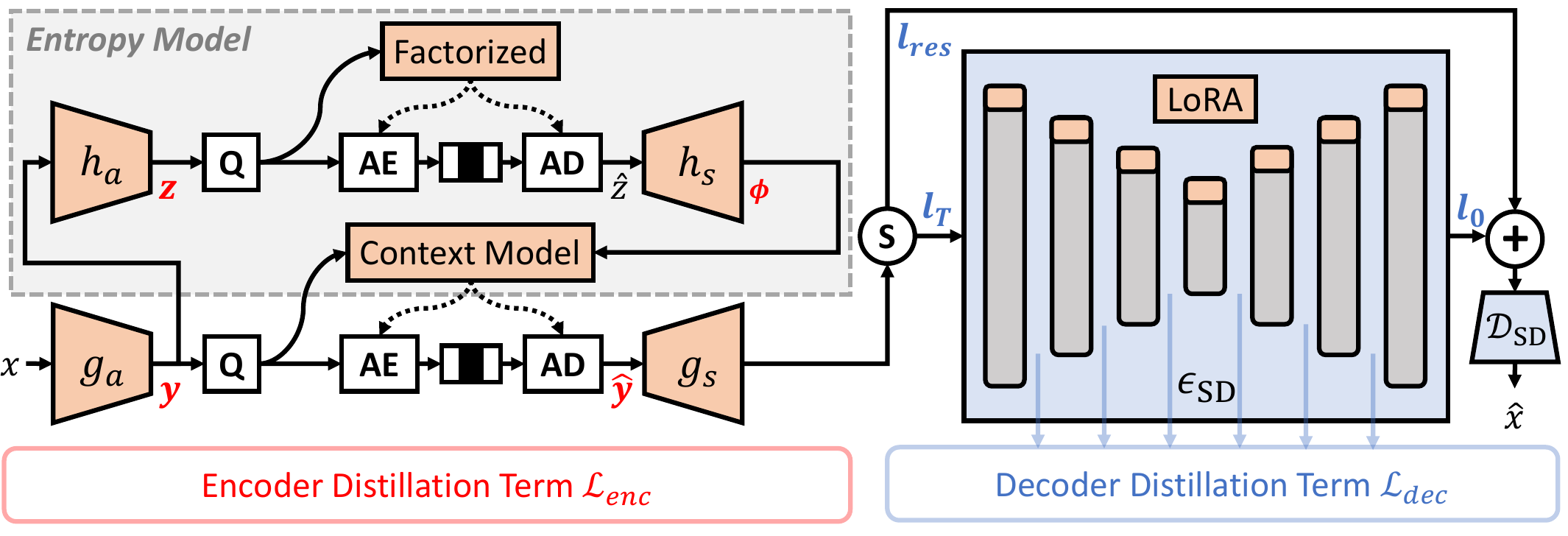}  
    \captionsetup{skip=2pt}
    \caption{Dual-side feature distillation from AEIC-ME to AEIC-SE using $\mathcal{L}_{enc}$ and $\mathcal{L}_{dec}$. We align intermediate encoder latents $y$, $z$, $\phi$ and $\hat{y}$ for the shallow encoder distillation, while facilitate $l_T$, $l_{res}$, $l_0$, and the internal Unet features for the decoder alignment.}
    \label{distill}
\end{figure}

\noindent\textbf{Dual-Side Feature Distillation.} As illustrated in Fig. \ref{distill}, the encoder-side distillation term $\mathcal{L}_{enc}$ leverages four intermediate latent representations: $y$, $z$, $\phi$, and $\hat{y}$, corresponding to the outputs of $g_a$, $h_a$, $h_s$, and the quantized $y$ as defined in Eq. \ref{eq2}, respectively. Due to the dimensional mismatch between the AEIC-ME (teacher) and AEIC-SE (student) encoders, we introduce a lightweight, trainable projection module $f(\cdot)$ implemented as a single convolutional layer to align feature spaces, formulating $\mathcal{L}_{enc}$ as:
\begin{equation}
\begin{aligned}
\mathcal{L}_{enc} &= \| y^{tea} - f(y^{stu}) \|_2^2 + \| z^{tea} - f(z^{stu}) \|_2^2 \\
&+ \| \phi^{tea} - f(\phi^{stu}) \|_2^2 + \| \hat{y}^{tea} - f(\hat{y}^{stu}) \|_2^2
\end{aligned}
\label{eq7}
\end{equation}

For the decoder-side supervision, the distillation term $\mathcal{L}_{dec}$ aligns both global and local reconstruction features. Specifically, we exploit the dual-branch decoding latents $l_T$ and $l_{res}$ (Eq. \ref{eq3}), the one-step denoising result $l_0$ (Eq.~\ref{eq4}), and the intermediate feature maps from the denoiser $\epsilon_{\mathrm{SD}}$. The decoder distillation loss is expressed as:
\begin{equation}
\begin{aligned}
\mathcal{L}_{dec} &= \| l_T^{tea} - f(l_T^{stu}) \|_2^2 + \| l_{res}^{tea} - f(l_{res}^{stu}) \|_2^2 \\
&+ \| l_0^{tea} - f(l_0^{stu}) \|_2^2 + \sum_n \| h_n^{tea} - f(h_n^{stu}) \|_2^2
\end{aligned}
\label{eq8}
\end{equation}
where $h_n$ denotes the feature representation from the $n$-th block of the UNet (down blocks, mid block and up blocks as defined in SD-Turbo \cite{sauer2024adversarial}). The projection function $f(\cdot)$ remains learnable to enable flexible feature alignment \cite{sunpocketsr}.

\subsection{Multi-Stage Progressive Training}
\label{sec:training}

\begin{figure*}[!h]
    \centering
    \includegraphics[width=\textwidth]{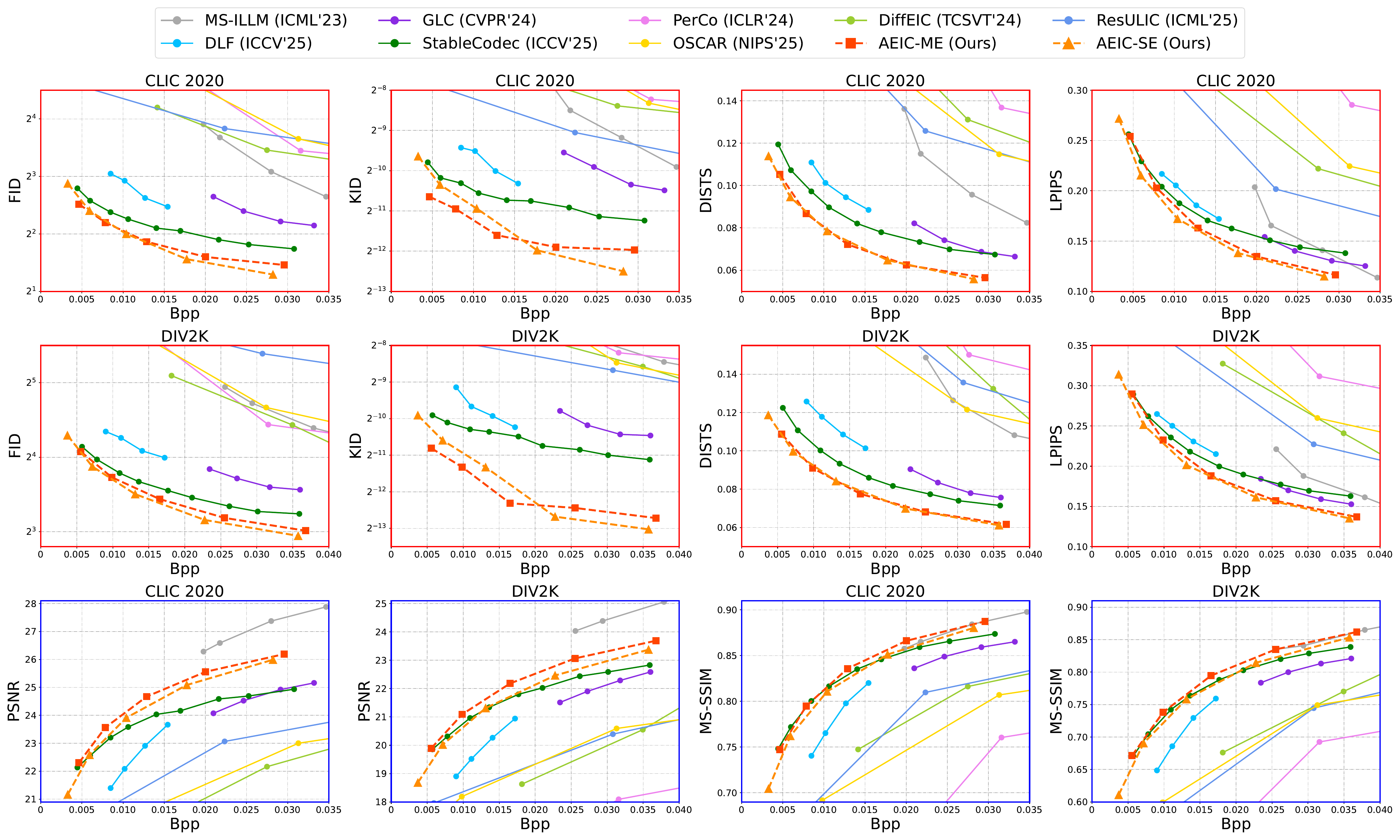}  
    \captionsetup{skip=2pt}
    \caption{Rate-perception (\textcolor{red}{sub-figures in red borders}) and rate-distortion (\textcolor{blue}{sub-figures in blue borders}) comparison of advanced ultra-low bitrate image compression methods on the CLIC 2020 test set and DIV2K validation set.}
    \label{rdpcurve}
\end{figure*}

In this section, we detail the progressive training for AEIC. We adopt end-to-end optimization joining bitrate and reconstruction constraints. Given the input-output image pair $(x, \hat{x})$, the quantized latent $\hat{y}$ and the quantized hyperprior latent $\hat{z}$ \cite{balle2018variational}, we construct our overall training objective based on the standard rate-distortion objective:
\begin{equation}
    \lambda \mathcal{R}(\hat{y}, \hat{z}) + \mathcal{D}(x, \hat{x})
\end{equation}
where the bitrate $\mathcal{R}$ and distortion $\mathcal{D}$ are balanced by the Lagrange multiplier $\lambda$, and the bitrate $\mathcal{R}$ is defined as:
\begin{equation}
    \mathcal{R}(\hat{y}, \hat{z})=\mathbb{E}\left[ -\log_2 p_{\hat{y}|\hat{z}}(\hat{y} \mid \hat{z}) \right] 
  + \mathbb{E}\left[ -\log_2 p_{\hat{z}}(\hat{z}) \right]
\end{equation}

\noindent \textbf{Teacher Learning. }We train AEIC-ME in two stages with different bitrate constraints $\lambda_{\mathrm{S1}}$ and $\lambda_{\mathrm{S2}}$ ($\lambda_{\mathrm{S1}} < \lambda_{\mathrm{S2}}$). In Stage 1, we adopt a relaxed bitrate constraint $\lambda_{\mathrm{S1}}$ to achieve stable end-to-end optimization. In Stage 2, we apply a larger $\lambda_{\mathrm{S2}}$ incorporated with a GAN loss to finetune towards ultra-low bitrates. The objectives for AEIC-ME are:
\begin{align}
\mathcal{L}_{\mathrm{S1}}^{tea}&=\lambda_{\mathrm{S1}}\mathcal{R}(\hat{y}, \hat{z})+\mathcal{D}(x, \hat{x}) \\
\mathcal{L}_{\mathrm{S2}}^{tea}&=\lambda_{\mathrm{S2}}\mathcal{R}(\hat{y}, \hat{z})+\mathcal{D}(x, \hat{x})+\alpha\mathcal{L}_{adv}
\end{align}
where
\begin{align}
\mathcal{D}(x, \hat{x})=\gamma_{1}||x-\hat{x}||_2^2+\gamma_{2}\mathcal{L}_{p}(x, \hat{x})+\gamma_{3}\mathcal{L}_{s}(x, \hat{x}) \\
\mathcal{L}_{adv} = 
\mathbb{E}\left[ \log_2 \mathbf{D}(x) \right]
+ \mathbb{E}\left[ \log_2(1 - \mathbf{D}(\mathbf{G}(x))) \right]
\end{align}
Here, the distortion term $\mathcal{D}$ comprises reconstruction loss, perceptual loss $\mathcal{L}_{p}$ \cite{zhang2018unreasonable} and semantic loss $\mathcal{L}_{s}$ \cite{lee2024neural}, while $\mathbf{D}$ and $\mathbf{G}$ denote the discriminator and the generator (namely our model), respectively. We switch $\mathcal{L}_{p}$ to overlap-chunked edge-aware DISTS \cite{wu2025omgsr, li2024unleashing} for better perceptual supervision under ultra-low bitrates in Stage 2. The discriminator is built upon DINOv3 \cite{simeoni2025dinov3} with trainable heads \cite{kumari2022ensembling}.

\noindent \textbf{Student Distillation. }As discussed in Section \ref{sec:distill}, we then distill AEIC-SE under the guidance of AEIC-ME. Specifically, we add an encoder distillation term $\mathcal{L}_{enc}$ to assist the shallow encoder training in student Stage 1, and guide the decoder convergence in student Stage 2 with the decoder distillation term $\mathcal{L}_{dec}$. The objectives for AEIC-SE are:
\begin{align}
\mathcal{L}_{\mathrm{S1}}^{stu}&=\lambda_{\mathrm{S1}}\mathcal{R}(\hat{y}, \hat{z})+\mathcal{D}(x, \hat{x})+\beta_1\mathcal{L}_{enc} \\
\mathcal{L}_{\mathrm{S2}}^{stu}&=\lambda_{\mathrm{S2}}\mathcal{R}(\hat{y}, \hat{z})+\mathcal{D}(x, \hat{x})+\alpha\mathcal{L}_{adv}+\beta_2\mathcal{L}_{dec}
\end{align}

\noindent\textbf{High-Resolution Finetuning.} We empirically observe that the shallow encoder tends to weaken the model’s generalization ability when applied to high-resolution images. To address this limitation, we introduce a short Stage 3 finetuning phase designed specifically for AEIC-SE, using large $1024\times1024$ image patches. The Stage 3 objective is formulated as follows, with re-initialized discriminator heads:

\begin{equation}
\mathcal{L}_{\mathrm{S3}}^{stu}=\lambda_{\mathrm{S3}}\mathcal{R}(\hat{y}, \hat{z})+\mathcal{D}(x, \hat{x})+\alpha\mathcal{L}_{adv}
\end{equation}

\section{Experiments}

\subsection{Implementation}

\textbf{Training Details.} Our training data consists of the training set of DF2K \cite{lim2017enhanced}, CLIC 2020 Professional \cite{toderici2020clic}, and the first 10K images from LSDIR \cite{li2023lsdir}. The data augmentation includes random horizontal and vertical flips. Stage 1 for AEIC takes over 300K iterations, using 512$\times$512 patches with a batch size of 8. In Stage 2, we use DF2K and CLIC 2020 to finetune AEIC for another 30K iterations with GAN incorporated, increasing the batch size to 32. Stage 3 for AEIC-SE takes only 5K iterations using 1024$\times$1024 patches with a batch size of 8. We set LoRA ranks to 32, and $\{\beta_1, \beta_2\}$ to $\{0.5, 0.001\}$. All models are trained using 2$\times$ RTX 3090 GPUs with PyTorch gradient accumulation and gradient checkpointing. Additional hyper parameters and training details are reported in the supplementary.

\noindent \textbf{Test Data.} We adopt the test set of CLIC 2020 Professional \cite{toderici2020clic} (CLIC 2020 Test), the validation set of DIV2K \cite{agustsson2017ntire} (DIV2K Val) and Kodak \cite{kodak} following \cite{jia2024generative, li2024towards, Xue_2025_ICCV, Zhang_2025_ICCV}. CLIC 2020 Test and DIV2K Val contain 428 and 100 high-quality images with 2K resolution, respectively, while Kodak contains 24 images with a smaller resolution of 768$\times$512. We adopt the tiling techniques \cite{Zhang_2025_ICCV, wang2024exploiting, zhang2024degradation} on the Unet and VAE decoder for high-resolution images. 

\begin{figure*}[!ht]
    \centering
    \includegraphics[width=\textwidth]{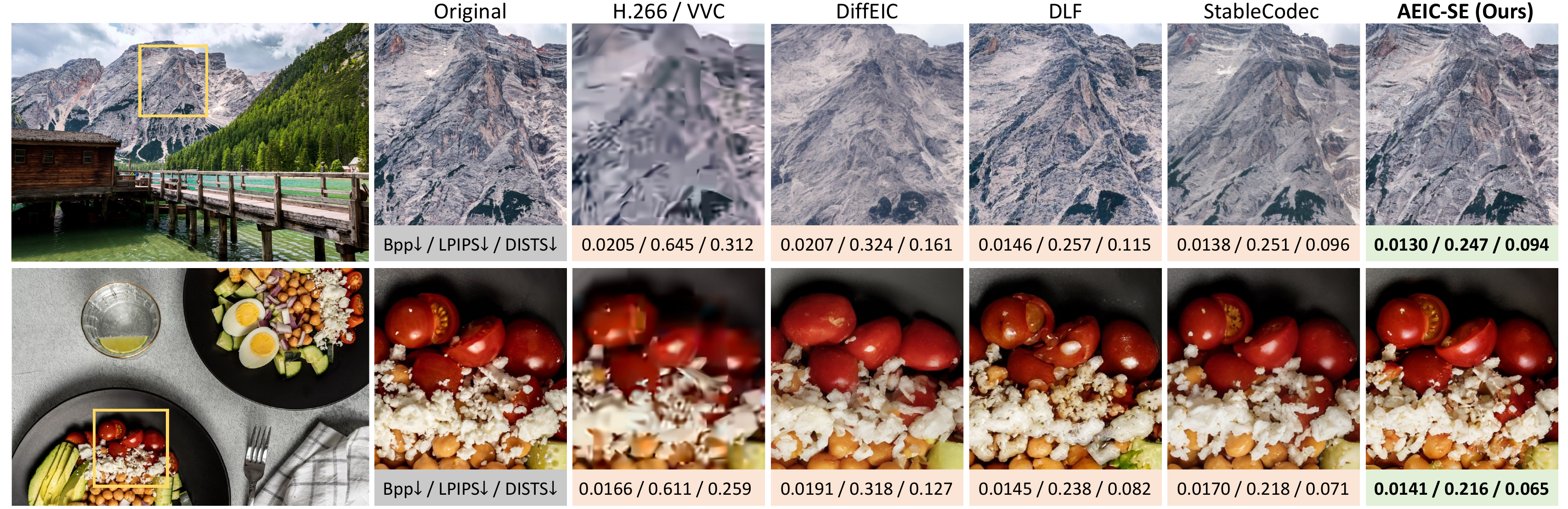}  
    \caption{Qualitative results on the CLIC 2020 test set. We use VTM-23.13 intra for H.266/VVC \cite{bross2021overview}. Best viewed on screen for details.}
    \label{visual}
\end{figure*}

\begin{table*}[!ht]
    \centering
    \setlength{\tabcolsep}{0.9mm}
    \caption{Practical coding latency (ms) on two kinds of GPUs and image resolutions. Both the encoding and decoding process include the autoregressive entropy coding with the entropy model. The best results are highlighted in \textbf{bold}, while the best results among ultra-low bitrate codec are \underline{underlined}. "OOM" means out of memory. We also report the \textcolor{red}{[encoding frames per second] in red} for AEIC models.}
    \label{time}
    \begin{threeparttable}
    \small
        \begin{tabular}{c|c|c|c|c|c|c|c|c|c|c}
            \Xhline{1.0pt}
            \multirow{3}{*}{Type} & \multirow{3}{*}{Method} & \multirow{3}{*}{Steps} & \multicolumn{4}{c|}{Encoding} & \multicolumn{4}{c}{Decoding} \\
            \cline{4-11}
            ~ & ~ & ~ & \multicolumn{2}{c|}{GTX 1080Ti} & \multicolumn{2}{c|}{RTX 4090D} & \multicolumn{2}{c|}{GTX 1080Ti} & \multicolumn{2}{c}{RTX 4090D} \\
            \cline{4-11}
            ~ & ~ & ~ & 768$\times$512 & 1920$\times$1088 & 768$\times$512 & 1920$\times$1088 & 768$\times$512 & 1920$\times$1088 & 768$\times$512 & 1920$\times$1088 \\
            \hline
            \hline
            Normal      & ELIC \cite{he2022elic}          & - & 302.0   & 1010.1    & 90.4  & 300.5 & 458.0 & 1648.9    & 171.1 & 465.3     \\
            Bitrate     & EVC-Small \cite{wang2023evc}      & - & 34.7    & 146.6     & 20.0  & 35.3  & \textbf{32.1}  & \textbf{120.5}     & \textbf{14.4}  & \textbf{33.6}      \\
            \hline
                        & PerCo \cite{careil2023towards}          & 20 & OOM        & OOM          & 245.2      & OOM      & OOM      & OOM          &  2841.6     & OOM          \\
                        & DiffEIC \cite{li2024towards}        & 50 & 1002.4    & OOM          &  153.7     &   1935.4    & 12742.0      & OOM          &   4785.1    & 50786.5          \\
            Ultra-Low   & DLF \cite{Xue_2025_ICCV}            & - & 493.9   & OOM       & 94.6  & 508.1 & 663.1 & OOM       & 152.1 & 962.5     \\
            Bitrate     & StableCodec \cite{Zhang_2025_ICCV}    & 1 & 328.2   & OOM       & 98.9  & 538.4 & 709.3 & OOM       & 192.9 & 1225.0    \\
                        & AEIC-ME (Ours)  & 1 & 60.3$\mathrm{_{\ \color{red}[16.6]}}$    & 284.0$\mathrm{_{\ \color{red}[3.5]}}$     & 37.1$\mathrm{_{\ \color{red}[27.0]}}$  & 58.7$\mathrm{_{\ \color{red}[17.0]}}$  & 433.4 & 3783.1    & 106.7 & 845.3     \\
                        & AEIC-SE (Ours)  & 1 & \underline{\textbf{22.6}}$\mathrm{_{\ \color{red}[44.2]}}$    & \underline{\textbf{110.4}}$\mathrm{_{\ \color{red}[9.1]}}$     & \underline{\textbf{14.0}}$\mathrm{_{\ \color{red}[71.4]}}$  & \underline{\textbf{27.9}}$\mathrm{_{\ \color{red}[35.8]}}$  & \underline{399.6} & \underline{3629.4}    & \underline{104.2} & \underline{836.2}     \\
            \Xhline{1.0pt}
        \end{tabular}
    \end{threeparttable}
\end{table*}

\begin{table}[!t]
    \centering
    \setlength{\tabcolsep}{1.1mm}
    \caption{Complexity comparison in parameters (M) and MACs (K) per pixel. The encoding and decoding both include the autoregressive entropy model. The best results are highlighted in \textbf{bold}, while the best results in ultra-low bitrate codec are \underline{underlined}. Note that ELIC and EVC-Small are normal bitrate compression methods.}
    \label{speed}
    \begin{threeparttable}
    \small
        \begin{tabular}{c|c|c|c|c|c}
            \Xhline{1.0pt}
            \multirow{2}{*}{Method} & \multirow{2}{*}{Steps} & \multicolumn{2}{c|}{Encoding} & \multicolumn{2}{c}{Decoding} \\
            \cline{3-6}
            ~ & ~ & Params. & MACs & Params. & MACs \\
            \hline
            \hline
            ELIC \cite{he2022elic}          & - & 33.38     & 346.03    & 33.38 & 590.823 \\
            EVC-Small \cite{wang2023evc}      & - & \textbf{11.64}     & 71.12     & \textbf{11.96} & \textbf{71.416}  \\
            \hline
            PerCo \cite{careil2023towards}      & 20 & $>$1B & 2666.47 & $>$1B & $>10^4$ \\
            DiffEIC \cite{li2024towards}    & 50 & 73.85 & 2253.69 & $>$1B & $>10^4$ \\
            DLF \cite{Xue_2025_ICCV}            & - & 437.35    & 1915.35   & \underline{561.63} & 4160.99 \\
            StableCodec \cite{Zhang_2025_ICCV}  & 1 & 102.23    & 2537.51   & 985.27 & 6201.66 \\
            AEIC-ME (Ours)        & 1 & 55.91     & 204.26    & 951.47 & 2884.88 \\
            AEIC-SE (Ours)     & 1 & \underline{16.10}     & \underline{\textbf{46.02}}     & 913.66 & \underline{2762.22} \\
            \Xhline{1.0pt}
        \end{tabular}
        \vspace{-1pt}
    \end{threeparttable}
\end{table}

\noindent \textbf{Evaluation.} We evaluate AEIC by the rate-distortion-perception and computational complexity performance. Concretely, we measure bitrate by bits per pixel (bpp), perceptual quality using FID \cite{heusel2017gans}, KID \cite{binkowski2018demystifying}, DISTS \cite{ding2020image} and LPIPS \cite{zhang2018unreasonable} (using AlexNet features by default), and distortion using PSNR and MS-SSIM \cite{wang2003multiscale}. We implement these metrics following \cite{Zhang_2025_ICCV}. For complexity, we report the practical coding latency, model parameters and MACs per pixel.

\subsection{Rate-Distortion-Perception Performance}

We compare the rate-distortion-perception performance of AEIC with advanced ultra-low bitrate codec MS-ILLM \cite{muckley2023improving}, GLC \cite{jia2024generative}, PerCo (SD) \cite{careil2023towards, korber2024perco}, DiffEIC \cite{li2024towards}, ResULIC \cite{keultra}, OSCAR \cite{guo2025oscar}, DLF \cite{Xue_2025_ICCV} and StableCodec \cite{Zhang_2025_ICCV}. 


Fig. \ref{rdpcurve} presents the rate-perception and rate-distortion curves comparison. Among all compared methods, AEIC demonstrates substantially better performance across all perceptual metrics, while maintaining competitive distortion results. Notably, AEIC-ME consistently outperforms StableCodec \cite{Zhang_2025_ICCV}, one of the latest extreme image codec, across the entire ultra-low bitrate range (0.005-0.035 bpp) in both rate-perception and rate-distortion performance. Furthermore, although AEIC-SE exhibits slightly worse distortion than AEIC-ME, it performs comparably or even better on perceptual metrics, indicating shallow encoder remains a viable and efficient solution for ultra-low bitrate perceptual image compression. Fig. \ref{visual} provides qualitative comparison on 2K resolution images. AEIC-SE produces more realistic and consistent details with fewer bits among all compared codec. More results are provided in the supplementary.

\subsection{Computational Complexity}

Owing to its shallow encoder, AEIC-SE demonstrates superior encoding efficiency. Table \ref{speed} compares model parameters and MACs per pixel among representative image compression methods. AEIC-SE achieves comparable encoding complexity to EVC-Small \cite{wang2023evc}, a highly efficient normal-bitrate codec designed for real-time applications, while being significantly more lightweight than other ultra-low bitrate codecs. Benefiting from the one-step diffusion and lite VAE decoder, AEIC-SE also exhibits reduced decoding computation relative to ultra-low bitrate approaches.  

To further assess practical efficiency, Table \ref{time} reports the encoding and decoding latency across methods. AEIC-SE attains the fastest encoding speed among all compared codecs while maintaining the best decoding speed within the ultra-low bitrate category. Specifically, AEIC-SE achieves real-time encoding at 35.8 frames per second (FPS) on 1080P images, delivering $18.2\times$ and $19.3\times$ speedups over DLF and StableCodec, respectively. Moreover, Both AEIC models support full-resolution 1080P encoding and decoding without tiling on a consumer-grade GTX 1080Ti GPU (11GB memory), underscoring its strong potential for source-limited practical encoding scenarios.

\section{Ablation Study}

\begin{table}[!t]
    \centering
    \setlength{\tabcolsep}{1.0mm}
    \caption{Ablation study on the spatial compression ratio. We stack downsample / upsample stages in $g_a$ / $g_s$ to construct AEIC-ME models with different spatial compression ratios. The rest of model remains the same. BD-rate is computed using four ultra-low bitrates among 0.01-0.035bpp. The anchor is StableCodec.}
    \label{ablation1}
    \small
    \begin{tabular}{c|cccc}
        \Xhline{1.0pt}
        \multirow{2}{*}{Method (Sp. Comp. Ratio)} & \multicolumn{4}{c}{BD-rate ($\downarrow$\%) on Kodak} \\
        \cline{2-5}
        ~ & PSNR & MS-SSIM & LPIPS & DISTS \\
        \hline
        \hline
        StableCodec (64$\times$) & 0 & 0 & 0 & 0 \\
        \hline
        AEIC-ME (64$\times$) & +14.30 & -4.54 & -6.97 & -6.95 \\
        AEIC-ME (32$\times$) & -2.21 & -4.85 & \textbf{-13.67} & \textbf{-24.91} \\
        AEIC-ME (16$\times$) & \textbf{-5.53} & \textbf{-8.00} & -1.91 & -9.03 \\
        \Xhline{1.0pt}
    \end{tabular}
\end{table}

\begin{table}[!t]
    \centering
    \caption{Ablation study on distillation. All models are trained fairly with the same strategy without HRF. BD-rate is computed using four ultra-low bitrates among 0.01-0.035bpp. Note that in this ablation, images in DIV2K are evaluated by 768$\times$512 patches.}
    \label{ablation2}
    \small
    \begin{tabular}{c|cc|ccc}
        \Xhline{1.0pt}
        \multirow{2}{*}{Model} & \multicolumn{2}{c|}{Distillation Term} & \multicolumn{3}{c}{BD-rate ($\downarrow$\%) on DIV2K} \\
        \cline{2-6}
        ~ & $\mathcal{L}_{enc}$ & $\mathcal{L}_{dec}$ & LPIPS & DISTS & FID \\
        \hline
        \hline
        AEIC-ME & - & - & 0 & 0 & 0 \\
        \hline
        \multirow{3}{*}{AEIC-SE} & - & - & +8.47 & +23.75 & +22.10 \\
        ~ & \checkmark & - & +3.92 & +7.68 & +4.75 \\
        ~ & \checkmark & \checkmark & \textbf{+0.60} & \textbf{+2.55} & \textbf{+2.98} \\
        \Xhline{1.0pt}
    \end{tabular}
\end{table}

\noindent \textbf{Spatial Compression Ratio for Extreme Bitrate.}  
Unlike adopting complex encoders and latent-space transform coding, AEIC employs a single custom encoder to learn a direct mapping from pixels to ultra-low bitrate latents. To investigate efficient encoder designs for extreme bitrates, we conduct an ablation on the spatial compression ratio, as summarized in Table \ref{ablation1}. Specifically, we vary the number of downsample and upsample stages in $g_a$ and $g_s$, respectively, to construct AEIC variants that produce latents $y$ of different spatial resolutions. While normal-bitrate codecs commonly adopt a $16\times$ spatial compression ratio, Table \ref{ablation1} suggests $32\times$ achieves a better balance between rate, distortion, and perception under ultra-low bitrates. In contrast, a shallower ratio ($16\times$) tends to overemphasize distortion metrics, whereas a more aggressive ratio ($64\times$) loses too much spatial information, resulting in degraded performance.  

\noindent \textbf{Knowledge Distillation for AEIC-SE.}  
Table \ref{ablation2} evaluates the dual-side feature distillation for AEIC-SE. Without $\mathcal{L}_{enc}$ and $\mathcal{L}_{dec}$, training AEIC-SE using the same protocol as AEIC-ME leads to severe performance degradation, as the shallow encoder struggles to learn expressive transforms without teacher's supervision. Incorporating encoder-side guidance $\mathcal{L}_{enc}$ significantly improves the efficiency of shallow encoder, particularly when evaluating reconstructions by perceptual metrics such as DISTS and FID. Furthermore, the decoder-side distillation term $\mathcal{L}_{dec}$ narrows the remaining performance gap between AEIC-ME and AEIC-SE to within $3\%$ on $768\times512$ images by aligning intermediate feature representations within the decoder.

\begin{figure}[!t]
    \centering
    \includegraphics[width=0.48\textwidth]{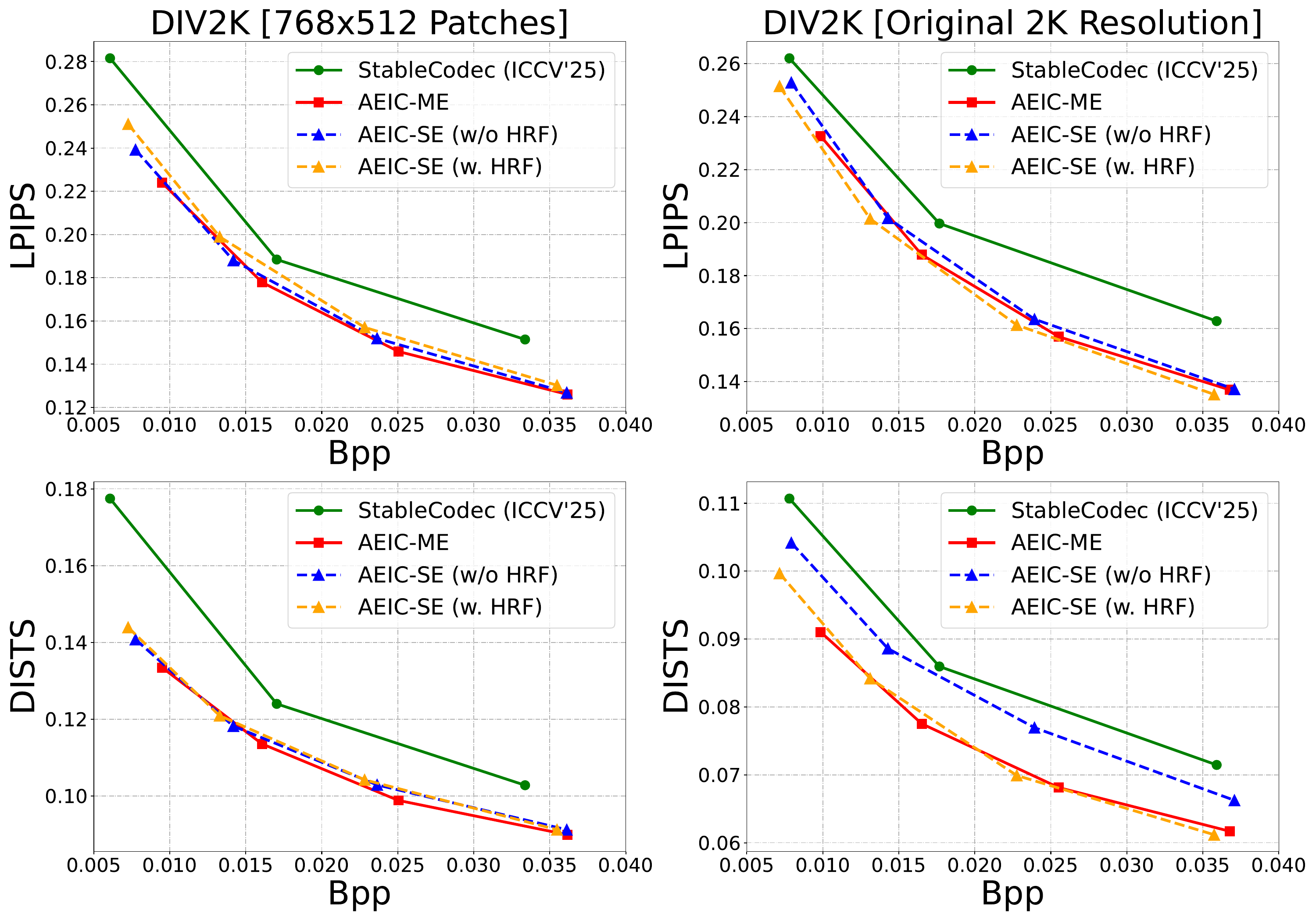}  
    \captionsetup{skip=2pt}
    \caption{Ablation study on the high-resolution finetuning (HRF) for AEIC-SE. We evaluate various methods on DIV2K by 512$\times$768 patches or full resolution. Note that StableCodec and our AEIC-ME are trained using patch size 512, while AEIC-SE adopts additional HRF with 5K iterations and patch size 1024.}
    \label{hrf}
\end{figure}

\noindent \textbf{High-Resolution Finetuning for AEIC-SE.}  
Shallow encoders typically exhibit limited generalization to large-resolution images (e.g., 1080P, 2K). We address this issue through a short high-resolution finetuning (HRF) stage. In Fig. \ref{hrf}, the performance of AEIC-SE, with and without HRF, is evaluated on DIV2K using both $768\times512$ patches and full 2K resolution inputs. On $768\times512$ patches, AEIC-SE performs comparably to AEIC-ME, and HRF introduces only a negligible bitrate shift with minimal impact on performance. However, at 2K resolution, HRF enhances AEIC-SE to a comparable performance to AEIC-ME and yields substantial gains particularly in DISTS evaluation.
\section{Conclusion}

In this work, we explored the feasibility of applying shallow encoder for extreme compression and introduced AEIC, an asymmetric extreme image compression framework that combines lightweight encoders with a one-step decoder. We further proposed a dual-side feature distillation strategy, effectively transferring knowledge from AEIC-ME to its shallow encoder variant AEIC-SE. Extensive experiments demonstrated that AEIC-SE achieves state-of-the-art perceptual quality and competitive distortion fidelity under ultra-low bitrates, while offering real-time 1080P encoding. We hope our findings underscores the potential of shallow encoder for ultra-low bitrate compression, and encourage practical extreme codec in source-limited applications.

{
    \small
    \bibliographystyle{ieeenat_fullname}
    \bibliography{main}
}

\clearpage

\setcounter{section}{0}
\maketitlesupplementary

\renewcommand\thesection{\Alph{section}}

\noindent This supplementary provides additional discussion on:
\begin{itemize}
\item Section \ref{A}. Further ablation study and discussion.
\item Section \ref{B}. Additional performance comparison.
\item Section \ref{C}. Study of user preference for AEIC-SE.
\item Section \ref{D}. Network structures of AEIC models.
\item Section \ref{E}. Detailed training and inference procedures.
\item Section \ref{F}. Third-party models and evaluation methods.
\item Section \ref{G}. Potential future directions based on AEIC-SE.
\end{itemize}

\section{Further Ablation and Discussion}
\label{A}

\begin{figure}[!ht]
    \centering
    \includegraphics[width=0.48\textwidth]{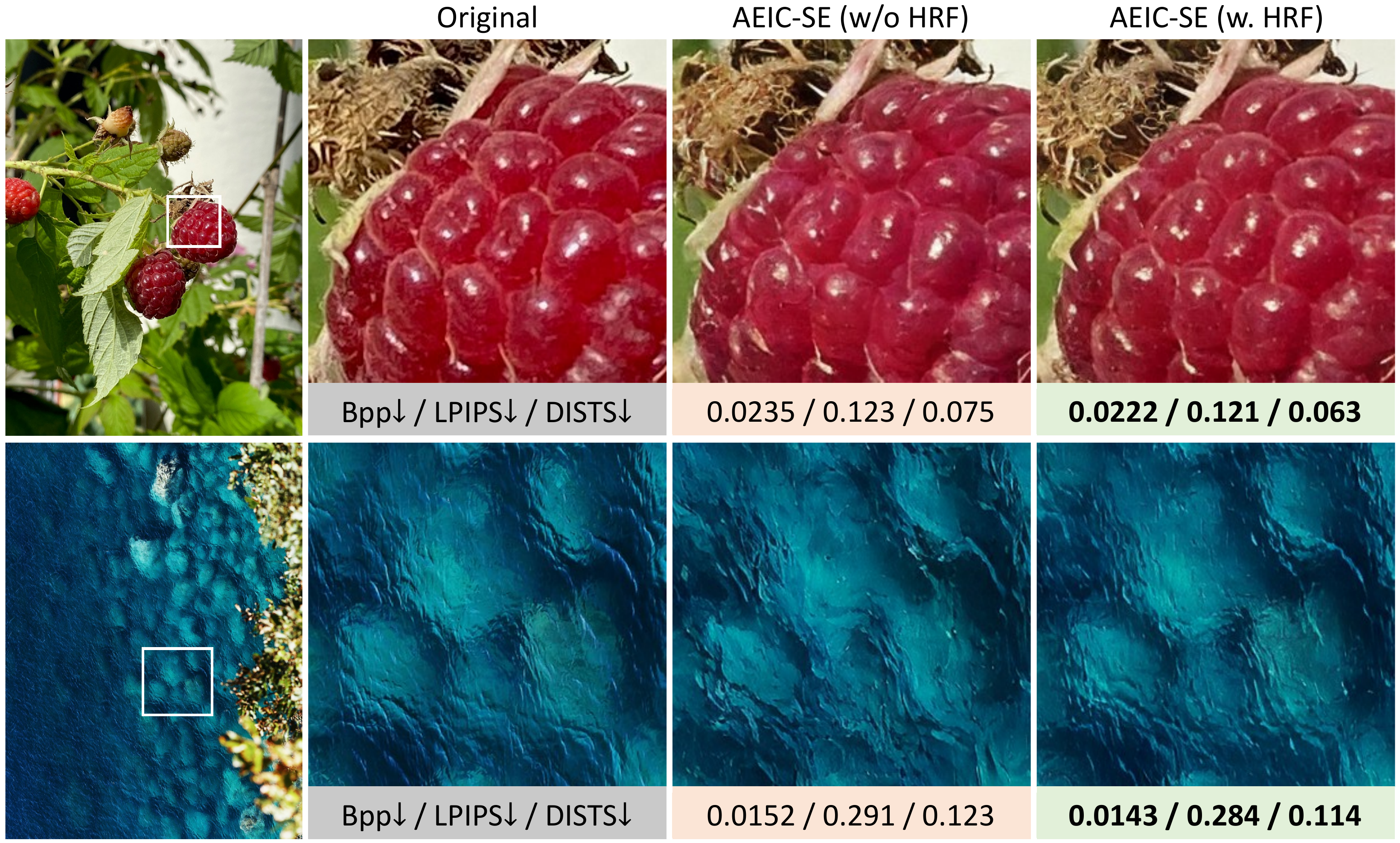}  
    \captionsetup{skip=2pt}
    \caption{Qualitative comparison on AEIC-SE models trained with or without high-resolution finetuning (HRF), using 2K resolution images from the CLIC 2020 test set. Best viewed on screen.}
    \label{hrf_visual}
    \vspace{-1em}
\end{figure}

\noindent \textbf{Effect of High-Resolution Finetuning.} As a supplement to Fig. 8 and its accompanying discussion, Fig. \ref{hrf_visual} provides a visual comparison of 2K-resolution reconstructions from AEIC-SE trained with and without high-resolution finetuning (HRF). After applying HRF, AEIC-SE produces reconstructions with more faithful and visually coherent local textures while using fewer bits. For example, the stem and contour of the berry become sharper and more consistent with the original content, whereas the water textures appear more realistic and better aligned with natural patterns.

\begin{figure}[!ht]
    \centering
    \includegraphics[width=0.48\textwidth]{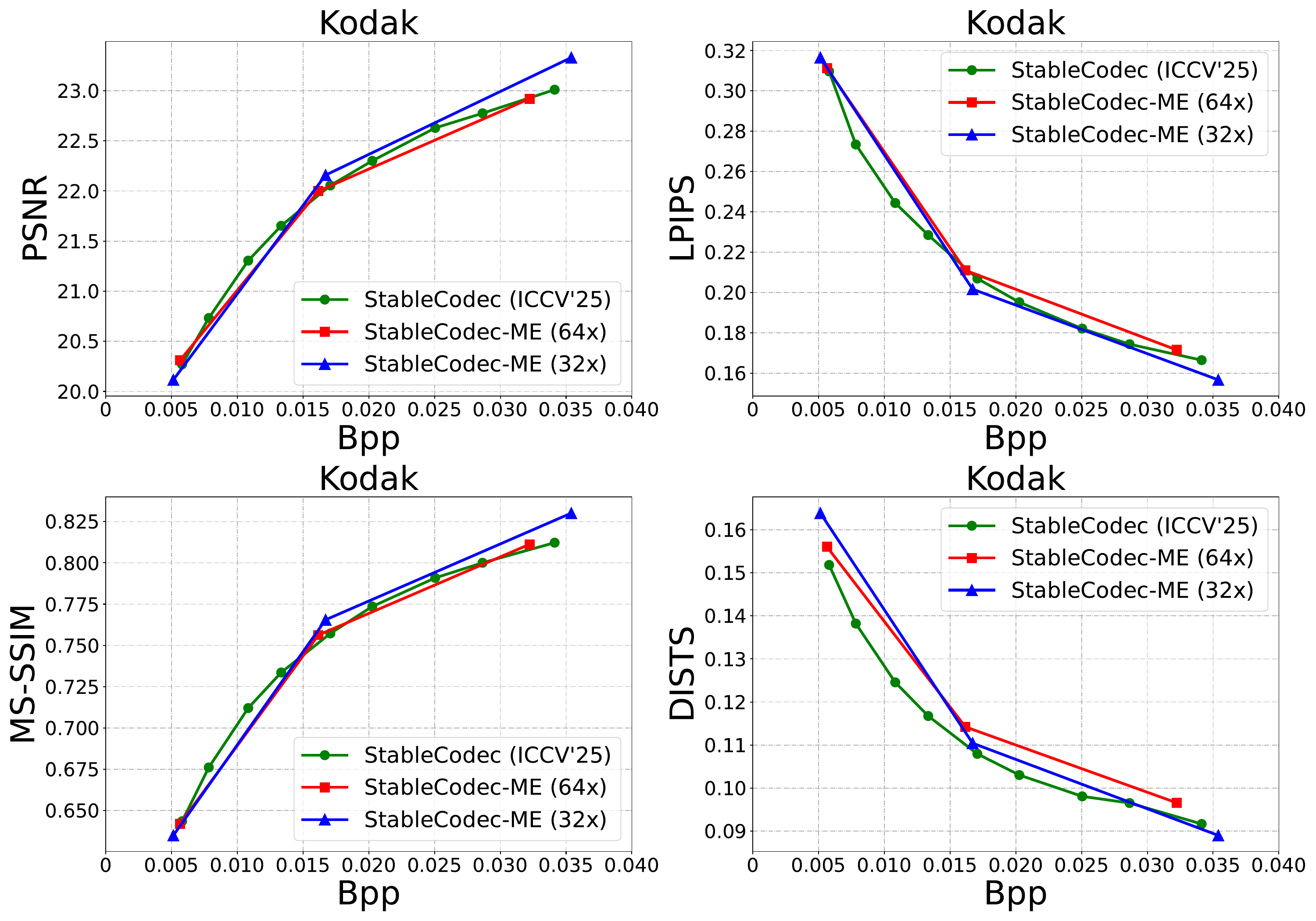}  
    \captionsetup{skip=2pt}
    \caption{StableCodec performance on Kodak when replacing original encoders with our moderate encoders (ME) of different spatial compression ratios (abbreviated as StableCodec-ME).}
    \label{supp_ablation1}
    \vspace{-1.25em}
\end{figure}

\noindent \textbf{Lightweight Encoder for Extreme Bitrate.} We further investigate whether lightweight encoders can be applied to StableCodec, one of the latest ultra-low bitrate image compression methods. StableCodec employs a complex multi-stage encoder that includes the Stable Diffusion VAE encoder, the ELIC encoder \cite{he2022elic}, and a latent-space transform encoder to produce a 64$\times$ downsampled latent. As shown in Fig. 3 (a), these components result in 47.16M encoder parameters in total. To examine the encoder complexity, we replace StableCodec's encoders with our moderate encoder (3.09M parameters) used in AEIC-ME, while keeping all other modules and training strategies unchanged. We test two variants with spatial compression ratios of 32 and 64, denoted as StableCodec-ME (32$\times$) and StableCodec-ME (64$\times$). As shown in Fig.~\ref{supp_ablation1}, StableCodec-ME (64$\times$) achieves performance comparable to the original StableCodec, whereas StableCodec-ME (32$\times$) even surpasses the baseline on all four metrics. These results support our finding that ultra-low bitrate compression does not require a large or expressive encoder, since the latent information is fundamentally constrained by the bitrate budget.

\noindent \textbf{Decoder Architectural Pruning.} We next provide a detailed analysis of the decoder architecture, specifically the unconditional denoiser $\epsilon_\mathrm{SD}$ and lite VAE decoder $\mathcal{D}_\mathrm{SD}$ introduced in Section 3.1.2. We begin by constructing a base AEIC-ME model using the original conditional denoiser $\epsilon_\mathrm{SD}$ and VAE decoder $\mathcal{D}_\mathrm{SD}$ from SD-Turbo. Following \cite{chen2025adversarial}, we remove the text encoder, timestep embeddings, and all cross-attention layers from $\epsilon_\mathrm{SD}$, since textual conditions contribute negligibly reconstruction quality in image compression \cite{vonderfecht2025lossy}, and the timestep input degenerates to a constant in one-step denoising. As shown in Table \ref{decoder_prune1} (Variant 1), this pruning removes over 75M parameters and converts $\epsilon_\mathrm{SD}$ into an unconditional denoiser (from Eq. 6 to Eq. 4), while also slightly improving overall performance.

StableCodec (Table 6) indicates that decoding latency is dominated by $\mathcal{D}_\mathrm{SD}$. To further improve decoding efficiency, we replace the original VAE decoder with a lite version \cite{chen2025adversarial} that prunes 50\% of its channels (Variant 2). Table \ref{decoder_prune1} shows that this reduces parameters from 49.5M to 12.4M, while incurring less than 1\% performance degradation relative to Variant 1. This efficiency-performance balance is reasonable because ultra-low bitrate compression (below 0.05 bpp) inherently cannot fully exploit the representational capacity of the original VAE decoder. Table \ref{decoder_prune2} further compares reconstruction performance across methods, indicating that even the highest bitrate setting of AEIC-ME produces reconstructions substantially worse than the SD VAE itself. Therefore, a lite decoder is sufficient for maintaining quality while enabling faster decoding.

\begin{table}[!t]
    \centering
    \setlength{\tabcolsep}{1.45mm}
    \caption{Ablation study on the decoder architecture pruning. We first construct a base AEIC-ME model with the original conditional denoiser $\epsilon_\mathrm{SD}$ and VAE decoder $\mathcal{D}_\mathrm{SD}$ in SD-Turbo. Then, we construct ``Variant 1" by removing text encoders, timestep embeddings, and all cross-attention layers from $\epsilon_\mathrm{SD}$, transforming $\epsilon_\mathrm{SD}$ into an unconditional denoiser. In ``Variant 2", we further replace the original $\mathcal{D}_\mathrm{SD}$ with a lite version \cite{chen2025adversarial} using only 50\% channels.}
    \vspace{-0.5em}
    \label{decoder_prune1}
    \small
    \begin{tabular}{c|cc|cccc}
        \Xhline{1.0pt}
        \multirow{2}{*}{Model} & \multicolumn{2}{c|}{Params. (M)} & \multicolumn{4}{c}{BD-rate ($\downarrow$\%) on Kodak} \\
        \cline{2-7}
        ~ & $\epsilon_\mathrm{SD}$ & $\mathcal{D}_\mathrm{SD}$ & PSNR & MS-SSIM & LPIPS & DISTS \\
        \hline
        \hline
        Base & 865.9 & 49.5 & 0 & 0 & 0 & 0 \\
        Variant 1 & \textbf{790.6} & 49.5 & \textbf{-1.29} & \textbf{-1.39} & \textbf{-0.23} & \textbf{-0.39} \\
        Variant 2 & \textbf{790.6} & \textbf{12.4} & -0.37 & -0.46 & +0.38 & +0.60 \\
        \Xhline{1.0pt}
    \end{tabular}
    \vspace{-0.5em}
\end{table}

\begin{table}[!t]
    \centering
    \setlength{\tabcolsep}{1.0mm}
    \caption{Reconstruction quality of different methods on Kodak. SD VAE stands for the VAE used in SD-Turbo and SD 2.1.}
    \vspace{-0.5em}
    \label{decoder_prune2}
    \small
    \begin{tabular}{c|cccc}
        \Xhline{1.0pt}
        Method & PSNR$\uparrow$ & MS-SSIM$\uparrow$ & LPIPS$\downarrow$ & DISTS$\downarrow$ \\
        \hline
        \hline
        SD VAE & \textbf{26.65} & \textbf{0.932} & \textbf{0.073} & \textbf{0.041} \\
        SD VAE (w. lite $\mathcal{D}_\mathrm{SD}$) & 26.56 & 0.930 & 0.079 & 0.046 \\
        AEIC-ME (0.038 bpp) & 23.30 & 0.832 & 0.143 & 0.082 \\
        \Xhline{1.0pt}
    \end{tabular}
    \vspace{-1em}
\end{table}

\noindent \textbf{Selection of Perceptual Loss.} Table \ref{oceadists} reports the impact of different perceptual losses when finetuning AEIC-ME under ultra-low bitrates. Unlike commonly adopted LPIPS, which measures latent-level distortion using VGG features, we employ DISTS \cite{ding2020image}, which imposes statistical constraints and provides more effective supervision for texture fidelity under extreme bitrates. In practice, we adopt the overlap-chunked edge-aware DISTS (OC-EA-DISTS) \cite{wu2025omgsr, li2024unleashing}, a recent variant tailored for different patch sizes and designed to jointly evaluate structure and texture similarity. As shown in Table \ref{oceadists}, using OC-EA-DISTS sacrifices distortion fidelity slightly but leads to improved perceptual quality, which is more critical in ultra-low bitrate scenarios.

\section{Additional Performance Comparison}
\label{B}

\noindent \textbf{Rate-Distortion-Perception Comparison on Kodak.} Fig. \ref{kodak} shows the rate-perception and rate-distortion comparisons on the Kodak dataset \cite{kodak}. Since Kodak contains only 24 images at a resolution of 768$\times$512, we follow prior works \cite{jia2024generative, Xue_2025_ICCV, Zhang_2025_ICCV} and omit FID \cite{heusel2017gans} and KID \cite{binkowski2018demystifying} due to their unreliability on small datasets. We compare AEIC with the traditional codec H.266/VVC \cite{bross2021overview} using VTM-23.13 intra mode, a distortion-oriented neural codec ELIC \cite{he2022elic}, and several state-of-the-art perceptual-oriented ultra-low bitrate methods including MS-ILLM \cite{muckley2023improving}, GLC \cite{jia2024generative}, PerCo \cite{careil2023towards}, DiffEIC \cite{li2024towards}, DLF \cite{Xue_2025_ICCV}, and StableCodec \cite{Zhang_2025_ICCV}. Both AEIC-ME and AEIC-SE achieve the best perceptual performance (e.g., LPIPS and DISTS) across all bitrates. In terms of distortion, AEIC-ME and AEIC-SE remain competitive among advanced perceptual-oriented codec, while significantly outperforming them in perception.

\begin{table}[!t]
    \centering
    \setlength{\tabcolsep}{1.0mm}
    \caption{Ablation study on the perceptual loss $\mathcal{L}_p$. We train AEIC-ME models using similar strategies as described in Section 3.3, only vary the selection of $\mathcal{L}_p$ in Stage 2 between LPIPS \cite{zhang2018unreasonable} and overlap-chunked edge-aware DISTS (OC-EA-DISTS) \cite{li2024unleashing, wu2025omgsr}.}
    \vspace{-0.5em}
    \label{oceadists}
    \small
    \begin{tabular}{c|c|cccc}
        \Xhline{1.0pt}
        \multirow{2}{*}{Model} & \multirow{2}{*}{$\mathcal{L}_p$ Selection} & \multicolumn{4}{c}{BD-rate ($\downarrow$\%) on Kodak} \\
        \cline{3-6}
        ~ & ~ & PSNR & MS-SSIM & LPIPS & DISTS \\
        \hline
        \hline
        \multirow{2}{*}{AEIC-ME} & LPIPS & \textbf{0} & \textbf{0} & 0 & 0 \\
        ~ & OC-EA-DISTS & +9.90 & +7.14 & \textbf{-5.19} & \textbf{-20.35} \\
        \Xhline{1.0pt}
    \end{tabular}
    \vspace{-1em}
\end{table}

\begin{figure}[!t]
    \centering
    \includegraphics[width=0.45\textwidth]{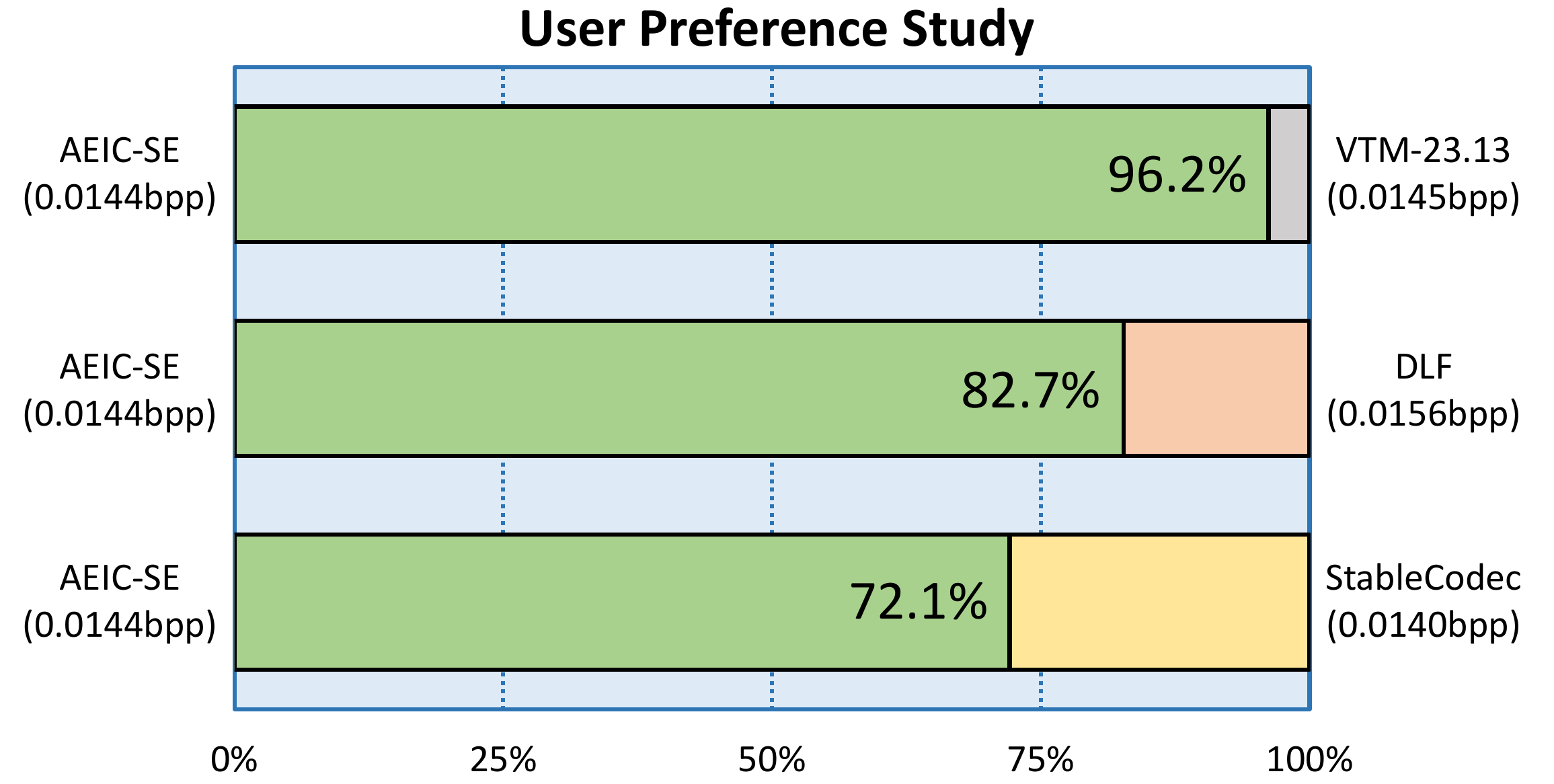}  
    \captionsetup{skip=2pt}
    \caption{User preference study on Kodak comparing AEIC-SE against traditional codec H.266/VVC and advanced learning-based generative codec DLF and StableCodec.}
    \label{us}
    \vspace{-1em}
\end{figure}

\noindent \textbf{Additional Visual Comparison.} Fig. \ref{visual1} presents additional qualitative results on 512$\times$512 patches from the Kodak dataset. We compare AEIC-SE with H.266/VVC (VTM-23.13 intra), as well as strong ultra-low bitrate baselines DLF and StableCodec. AEIC-SE consistently reconstructs more visually coherent structures and textures using fewer bits. Figs. \ref{visual2}-\ref{visual7} further provide comparisons on 2K-resolution images from the CLIC 2020 test set and DIV2K validation set. Across all resolutions and content types, AEIC-SE delivers the most visually consistent results while operating at the lowest bitrate, reinforcing its superior capability for ultra-low bitrate perceptual compression.

\section{User Study}
\label{C}

We conducted a user preference study based on side-by-side visual comparisons. In each case, we display the ground-truth image and two reconstructions at similar ultra-low bitrates: one produced by AEIC-SE and the other produced by a competitor (H.266/VVC, DLF and StableCodec). The left-right order of the two reconstruction methods was randomized to prevent positional bias. We invited 15 users. Each participant evaluated 24 cases. Fig. \ref{us} shows a clear preference, where AEIC-SE received 96.2\% of the votes against H.266/VVC, 82.7\% against DLF and 72.1\% against StableCodec, indicating consistently better visual quality.

\section{Model Structure}
\label{D}

The overall structure of AEIC models are detailed in Fig. \ref{network1} and Fig. \ref{network2}. Our codec consists of an analysis transform $g_a$, a synthesis transform $g_s$, and an entropy model. We follow \cite{li2023neural, Zhang_2025_ICCV} to construct our entropy model with a pair of hyper transform \cite{balle2018variational} and a 4-step quadtree-partitioned autoregressive context model. The major networks and hidden dimensions are detailed in Fig. \ref{network2}, exploiting efficient convolution blocks \cite{ma2024rewrite, yu2024mambaout, yu2024inceptionnext}. The synthesis transform $g_s$ produces two latents, $l_T$ and $l_{res}$, following the dual-branch decoding format \cite{Zhang_2025_ICCV}. Note that AEIC-ME and AEIC-SE only differ in the analysis transform $g_a$ and the entropy model as detailed in Fig. \ref{network2} and summarized in Table 1. Regarding the one-step diffusion, we set the LoRA rank in the unconditional Unet denoiser $\epsilon_\mathrm{SD}$ to 32, while keeping the pretrained VAE decoder $\mathcal{D}_\mathrm{SD}$ \cite{chen2025adversarial} frozen throughout AEIC training.

\section{Training and Inference Details}
\label{E}

\noindent \textbf{AEIC-ME training.} Stage 1 for AEIC-ME takes over 300K iterations, using 512$\times$512 patches and a batch size of 8. On 2$\times$ RTX 3090 GPUs (24GB memory), this process requires 4 gradient accumulation steps and an actual batch size of 1 for each GPU. The learning rate degrades from $1e^{-4}$ to $5e^{-5}$ after 280K iterations. $\{\lambda_{\mathrm{S1}}, \gamma_{1}, \gamma_{2}, \gamma_{3}\}$ are set to $\{1, 2, 1, 0.1\}$, respectively. Stage 2 takes over 30K iterations, increasing the batch size to 32. The learning rate starts from $5e^{-5}$, then degrades to $\{2e^{-5}, 1e^{-5}, 5e^{-6}, 1e^{-6}\}$ at $\{10, 25, 28, 29\}$K iterations. $\lambda_{\mathrm{S2}}$ chooses from $\{2, 4, 8, 16, 32\}$ for different ultra-low bitrates. $\alpha$ is set to 0.1. The total training for AEIC-ME requires approximately 9 days on 2$\times$ RTX 3090 GPUs.

\noindent \textbf{AEIC-SE training.} Stage 1 for AEIC-SE takes over 200K iterations, using 512$\times$512 patches and a batch size of 8. This process requires 2 gradient accumulation steps and an actual batch size of 2 for each GPU. The learning rate degrades from $1e^{-4}$ to $5e^{-5}$ after 180K iterations. $\{\lambda_{\mathrm{S1}}, \gamma_{1}, \gamma_{2}, \gamma_{3}, \beta_1\}$ are set to $\{1, 2, 1, 0.1, 0.5\}$, respectively. After 180K iterations, we drop $\mathcal{L}_{enc}$ and reset $\{\lambda_{\mathrm{S1}}, \gamma_{1}\}$ to $\{1.1, 0.5\}$ for fast convergence. Stage 2 takes over 30K iterations, using 512$\times$512 patches and a batch size of 32. The actual batch size and gradient accumulation step for each GPU are set to 1 and 16. The learning rate starts from $5e^{-5}$, then degrades to $\{2e^{-5}, 1e^{-5}, 5e^{-6}, 1e^{-6}\}$ at $\{10, 25, 28, 29\}$K iterations. $\lambda_{\mathrm{S2}}$ chooses from $\{2, 4, 8, 16, 32\}$ for different ultra-low bitrates. $\{\gamma_{1}, \gamma_{2}, \gamma_{3}, \alpha, \beta_2\}$ are set to $\{0.5, 1, 0.05, 0.05, 0.001\}$, respectively. After 20K iterations, we drop $\mathcal{L}_{dec}$ for fast convergence. Stage 3 for AEIC-SE takes over 5K iterations, using 1024$\times$1024 patches and a batch size of 8. The actual batch size and gradient accumulation step for each GPU are set to 1 and 4. Gradient checkpointing is activated. The learning rate starts from $2e^{-5}$, then degrades to $\{1e^{-5}, 5e^{-6}, 1e^{-6}\}$ at $\{3, 4.5, 4.8\}$K iterations. $\{\lambda_{\mathrm{S3}}, \gamma_{1}, \gamma_{2}, \gamma_{3}, \alpha\}$ remains the same as Stage 2. The total training for AEIC-SE also requires about 9 days.

\noindent \textbf{Inference Strategy.} We use similar tiling and color fix strategies \cite{Zhang_2025_ICCV, wang2024exploiting, zhang2024degradation} for high-resolution images. Specifically, for AEIC-ME we set Unet tile size to 96 with an overlap of 32, and the VAE decoder tile size to 160. Since AEIC-SE has been finetuned on 1024$\times$1024 patches, we set Unet tile size to 192 with an overlap of 64. 16-bit color fix \cite{Zhang_2025_ICCV} is employed when using tiling strategies for inference.

\section{Third-Party Models and Evaluation}
\label{F}

\noindent \textbf{Ultra-low Bitrate Image Codec.} We evaluate GLC \cite{jia2024generative}, DiffEIC \cite{li2024towards}, ResULIC \cite{keultra}, OSCAR \cite{guo2025oscar}, DLF \cite{Xue_2025_ICCV} and StableCodec \cite{Zhang_2025_ICCV} using the official code and pretrained weights. We finetune MS-ILLM \cite{muckley2023improving} using the official code and the pretrained weight (at the lowest available bitrate) to reach ultra-low bitrates. For PerCo \cite{careil2023towards}, we rely on a community implementation \cite{korber2024perco} and the pretrained weights as the official code is not available.

\noindent \textbf{Distortion-Oriented Neural Image Codec.} We evaluate EVC-Small \cite{wang2023evc} by the official code and pretrained weights. For ELIC \cite{he2022elic}, we follow the implementation in CompressAI \cite{begaint2020compressai}, and train models for ultra-low bitrates.

\noindent \textbf{Traditional Codec.} VTM-23.13 is the reference software for H.266/VVC \cite{bross2021overview}. We install the software according to the official instructions. For RGB images, we manage RGB-YUV420 transformation using FFmpeg as \cite{li2023neural}.

\noindent \textbf{Implementation of Evaluation Metrics.} We construct PSNR, MS-SSIM \cite{wang2003multiscale} and DISTS \cite{ding2020image} metrics using PyIQA with default settings, while implement LPIPS \cite{zhang2018unreasonable}, FID \cite{heusel2017gans} and KID \cite{binkowski2018demystifying} metrics using TorchMetrics. FID and KID are evaluated by splitting images into overlapped 256$\times$256 patches following the protocol in \cite{muckley2023improving}.

\section{Future Work}
\label{G}

In this work, we primarily focus on the feasibility of applying shallow encoder for source-limited ultra-low bitrate image compression senders. While the proposed AEIC-SE demonstrates strong perceptual quality, real-time practical encoding efficiency, and competitive decoding speed at ultra-low bitrates, a key challenge lies in further reducing the decoding latency, since achieving truly real-time decoding at extreme bitrates remains difficult due to the computational overhead of generative reconstruction. Future research may investigate more compact generative priors, hardware-friendly decoder designs, and novel decoder pruning mechanisms that preserve perceptual fidelity while significantly lowering computational costs. We hope these directions will inspire continued advancement toward efficient, deployable, and perceptually optimized ultra-low bitrate image compression systems.

\begin{figure*}[!h]
    \centering
    \includegraphics[width=\textwidth]{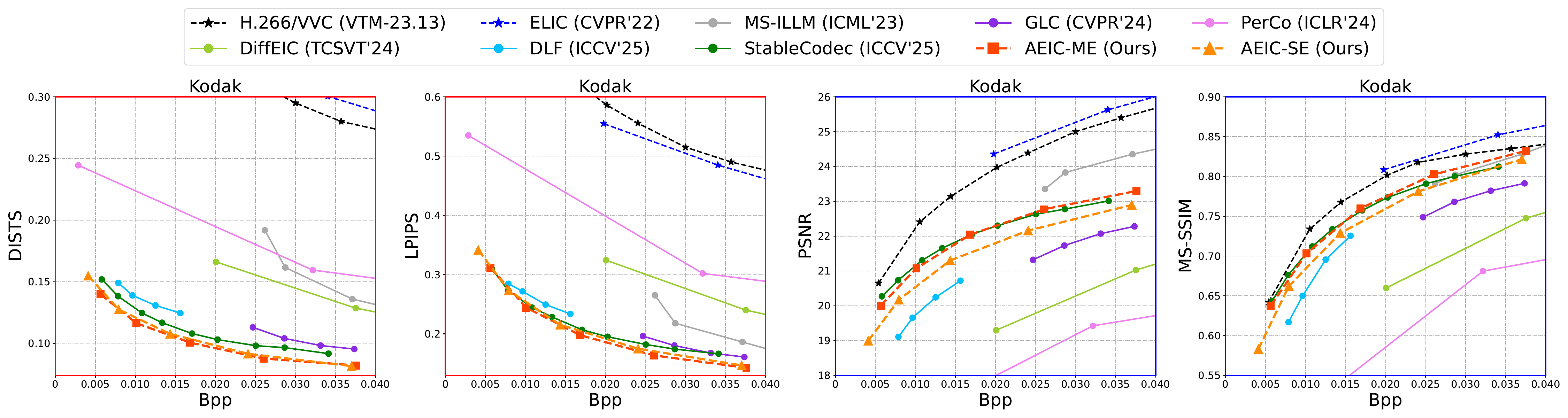}  
    \captionsetup{skip=2pt}
    \caption{Rate-perception (\textcolor{red}{sub-figures in red borders}) and rate-distortion (\textcolor{blue}{sub-figures in blue borders}) comparison of advanced image compression methods on the Kodak \cite{kodak} dataset. Note that FID and KID are not reported on Kodak due to its small size (24 images).}
    \label{kodak}
\end{figure*}

\begin{figure*}[!h]
    \centering
    \includegraphics[width=\textwidth]{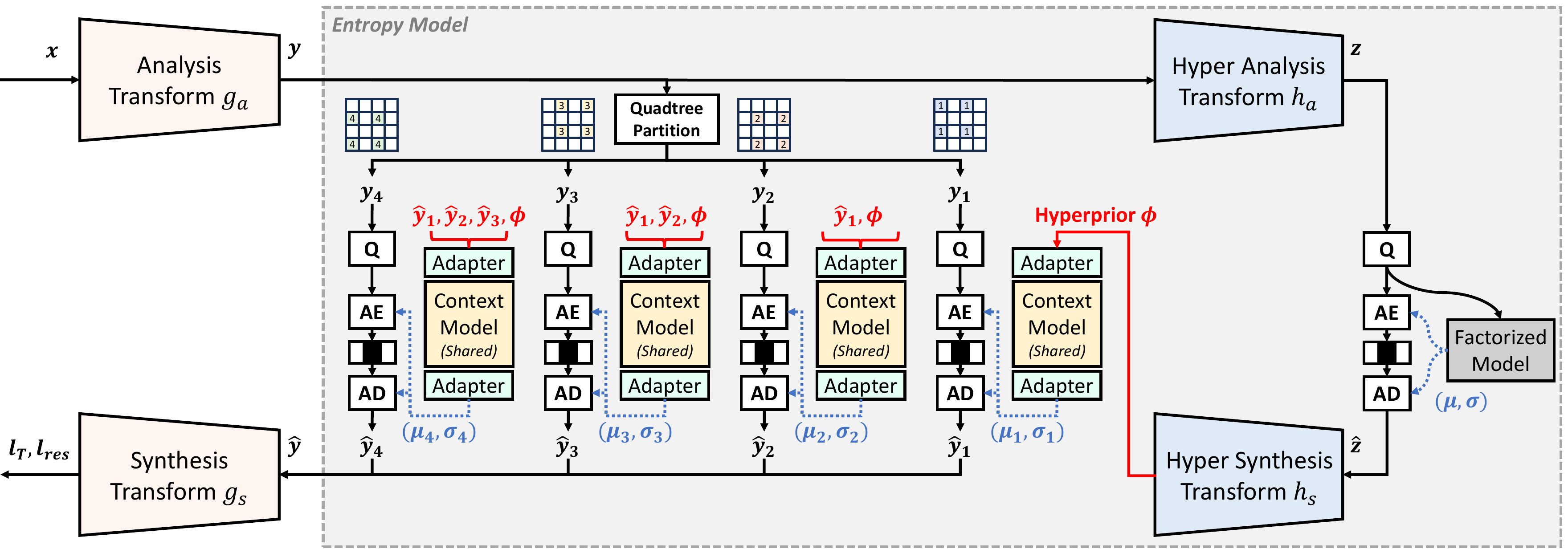}  
    \captionsetup{skip=2pt}
    \caption{Network structure of AEIC models. We follow \cite{minnen2018joint} to build our entropy model with a hyperprior \cite{balle2018variational} and an efficient 4-step autoregressive context model, exploiting quadtree partition \cite{li2023neural} and expressive networks \cite{Zhang_2025_ICCV}. Our moderate encoder variant (AEIC-ME) and shallow encoder variant (AEIC-SE) mainly differ in the analysis transform $g_a$ and the entropy model, as illustrated in Fig. \ref{network2}.}
    \label{network1}
\end{figure*}

\begin{figure*}[!h]
    \centering
    \includegraphics[width=\textwidth]{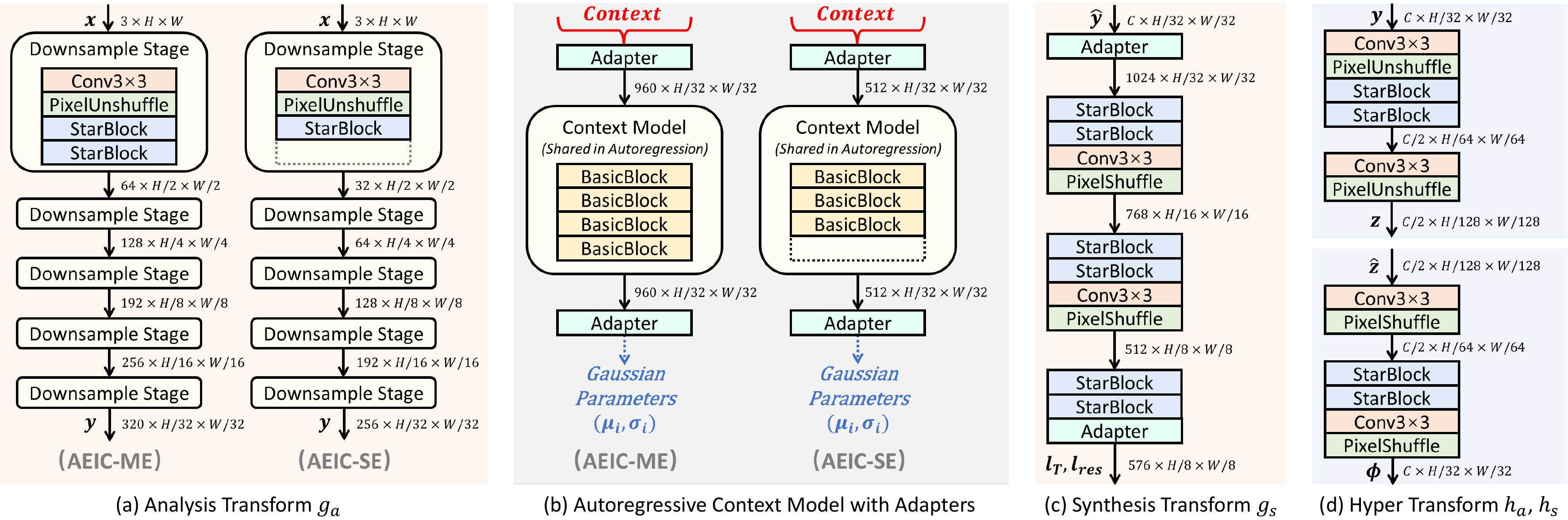}  
    \captionsetup{skip=2pt}
    \caption{Modules in AEIC models. Specifically, our transforms ($g_a$, $g_s$, $h_a$ and $h_a$) employ StarBlock \cite{ma2024rewrite}, an efficient convolution network based on element-wise multiplication. The context model follows a similar implementation of StableCodec with shared BasicBlocks (consists of InceptionNeXt \cite{yu2024inceptionnext} and GatedCNN \cite{yu2024mambaout}) and independent Adapters (a single resblock to adjust channel dimensions).}
    \label{network2}
\end{figure*}

\begin{figure*}[!h]
    \centering
    \includegraphics[width=\textwidth]{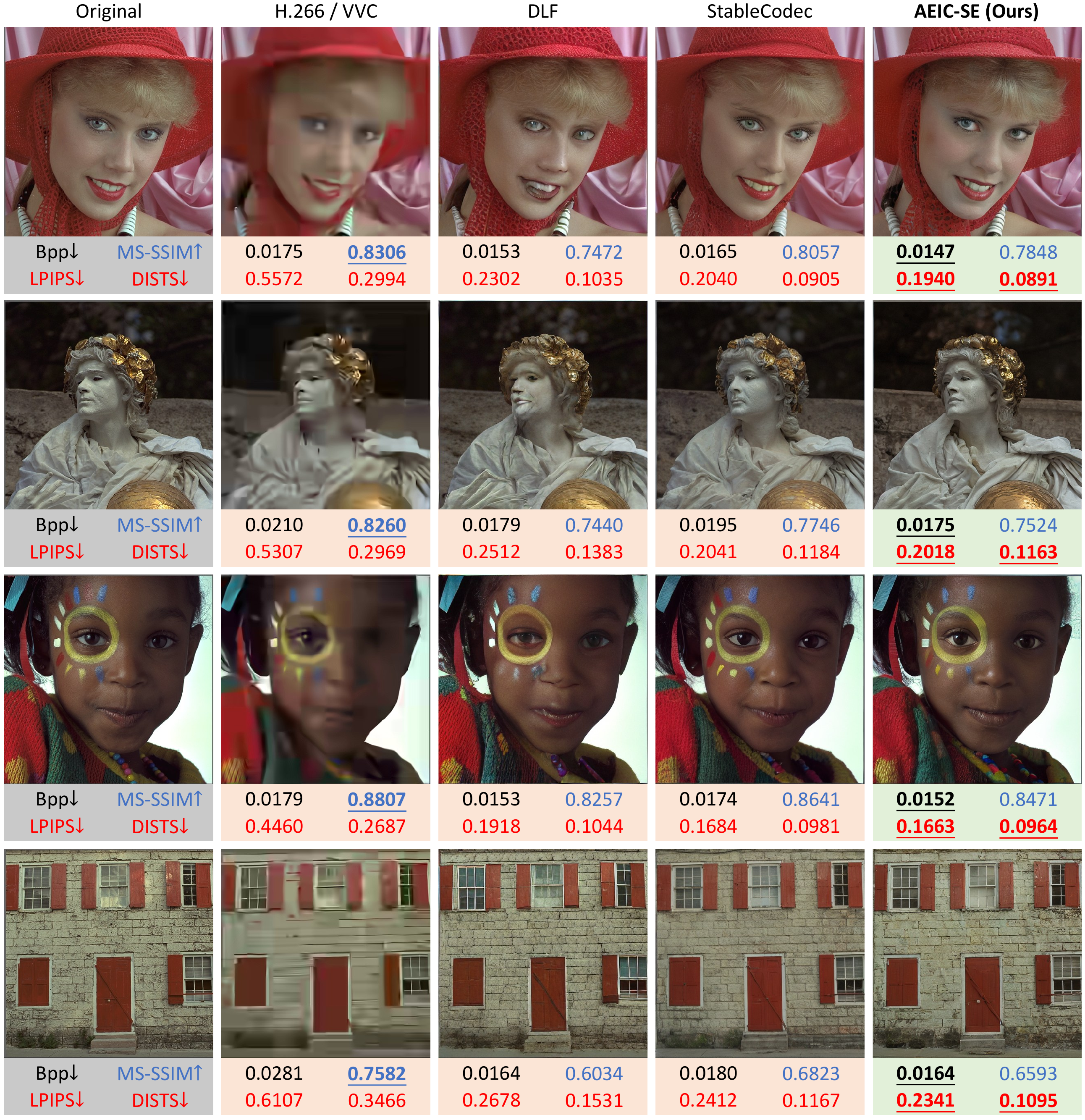}  
    \captionsetup{skip=2pt}
    \caption{Qualitative comparison (512$\times$512 patches) on the Kodak dataset. Distortion is evaluated with MS-SSIM, while perceptual quality is assessed using LPIPS and DISTS. The best results are highlighted in \textbf{\underline{bold and underlined}}. AEIC-SE achieves superior perceptual reconstruction with the fewest bits. Although H.266/VVC attains the highest MS-SSIM scores, its outputs exhibit blurriness and blocking artifacts, indicating that distortion metrics like MS-SSIM become less reliable at ultra-low bitrates.}
    \label{visual1}
\end{figure*}

\begin{figure*}[!h]
    \centering
    \includegraphics[width=\textwidth]{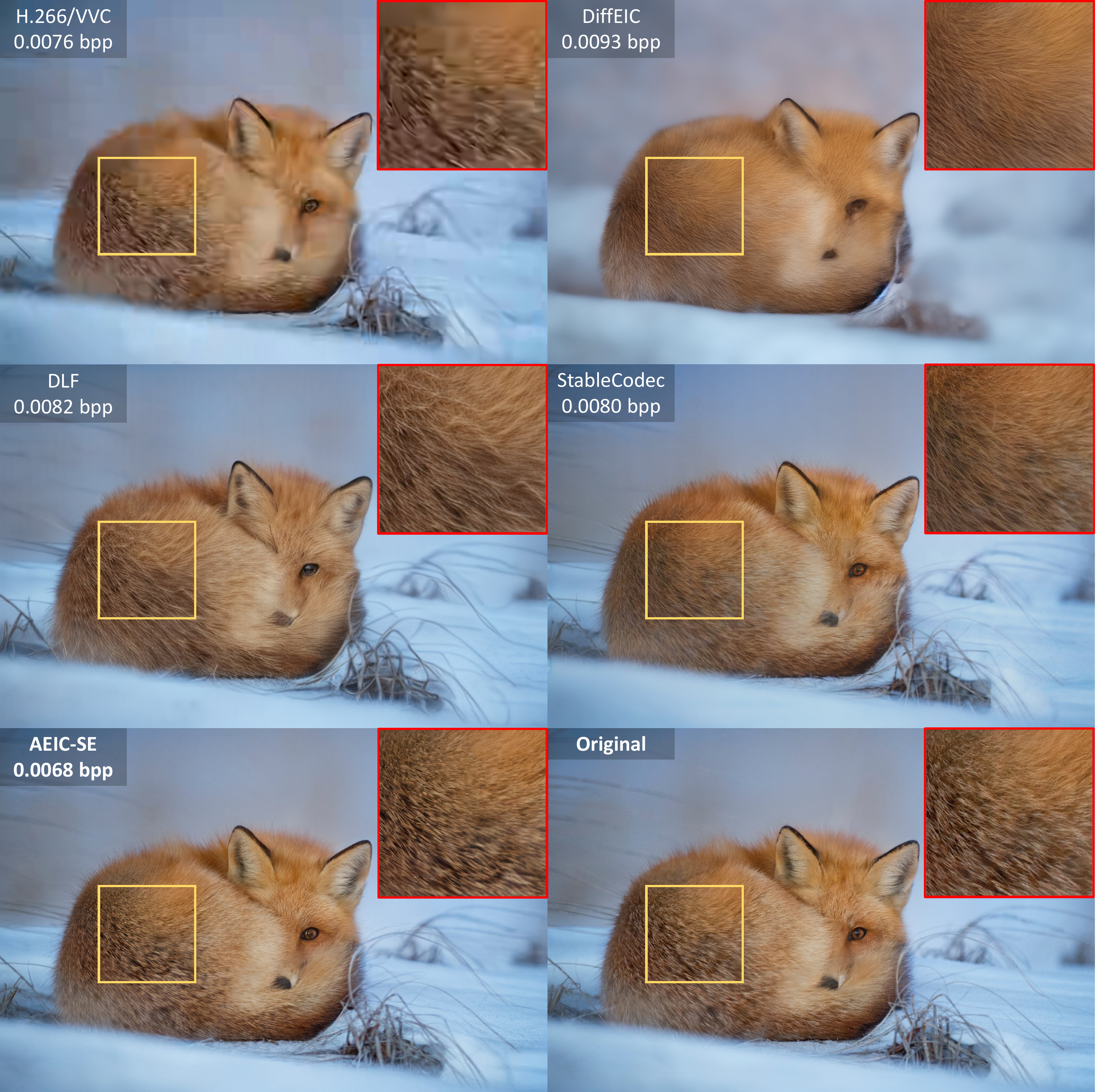}  
    \captionsetup{skip=2pt}
    \caption{Visual comparison (2K resolution) on the DIV2K validation set.}
    \label{visual2}
\end{figure*}

\begin{figure*}[!h]
    \centering
    \includegraphics[width=\textwidth]{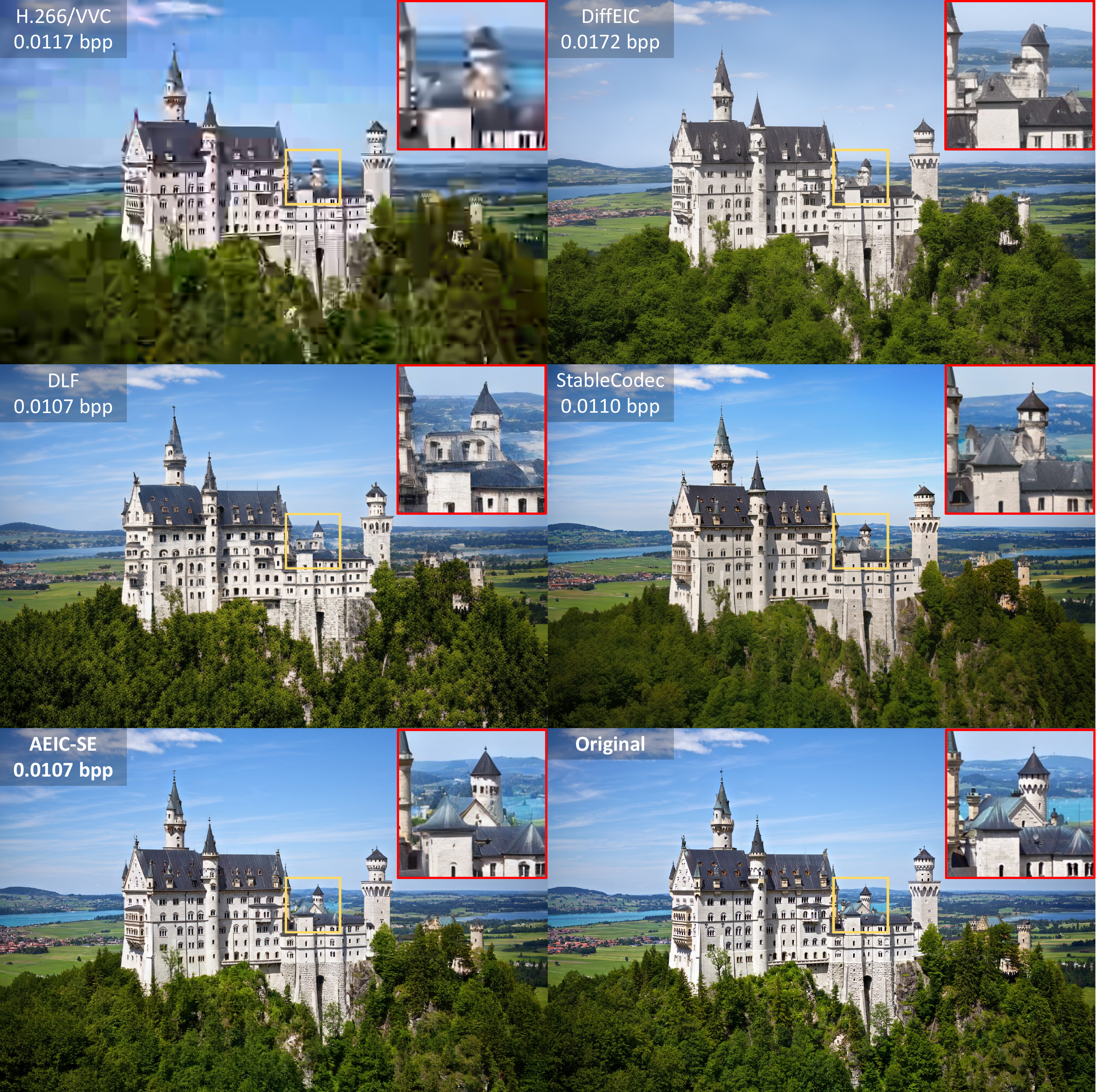}  
    \captionsetup{skip=2pt}
    \caption{Visual comparison (2K resolution) on the DIV2K validation set.}
    \label{visual3}
\end{figure*}

\begin{figure*}[!h]
    \centering
    \includegraphics[width=\textwidth]{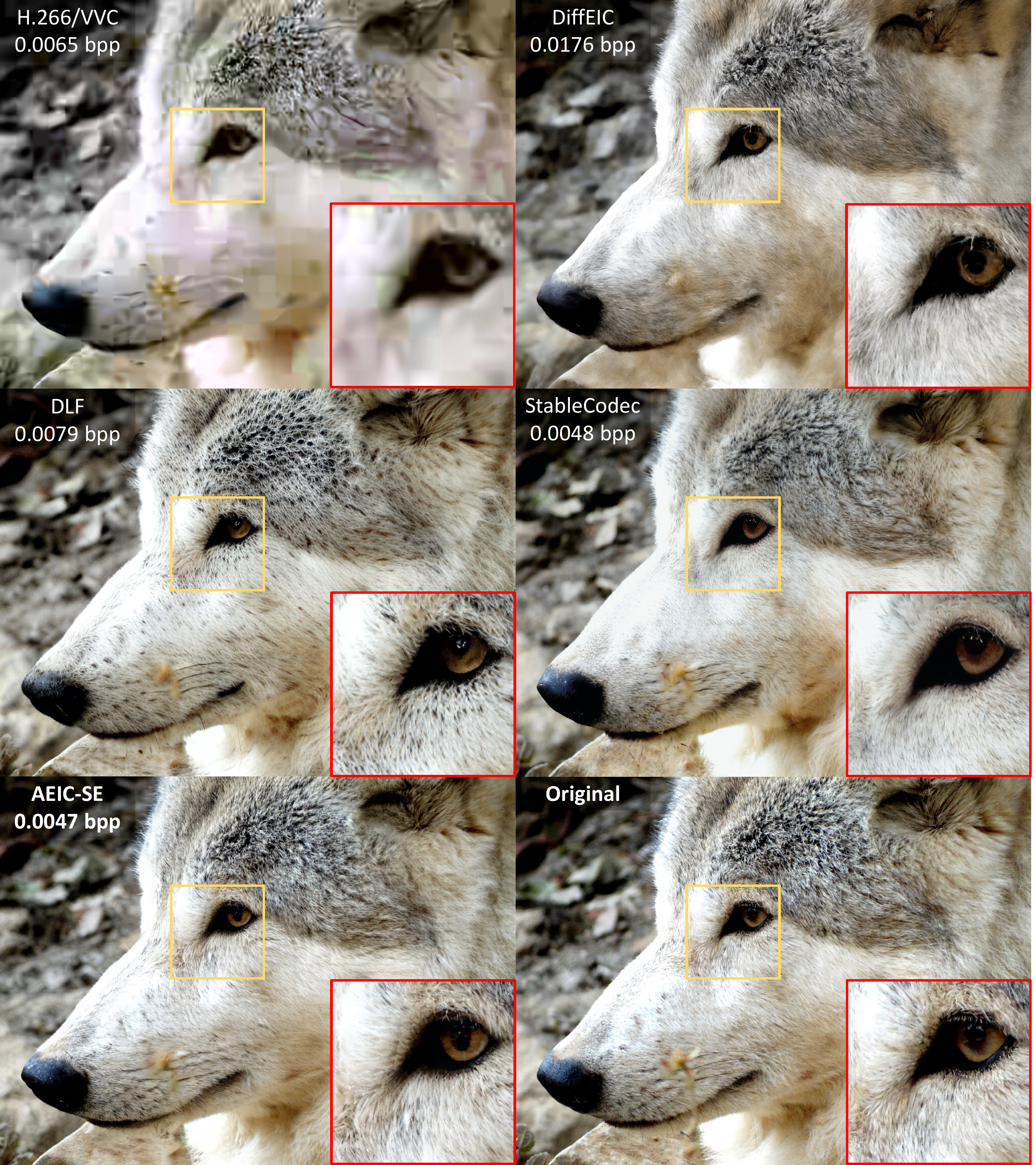}  
    \captionsetup{skip=2pt}
    \caption{Visual comparison (2K resolution) on the DIV2K validation set.}
    \label{visual4}
\end{figure*}

\begin{figure*}[!h]
    \centering
    \includegraphics[width=\textwidth]{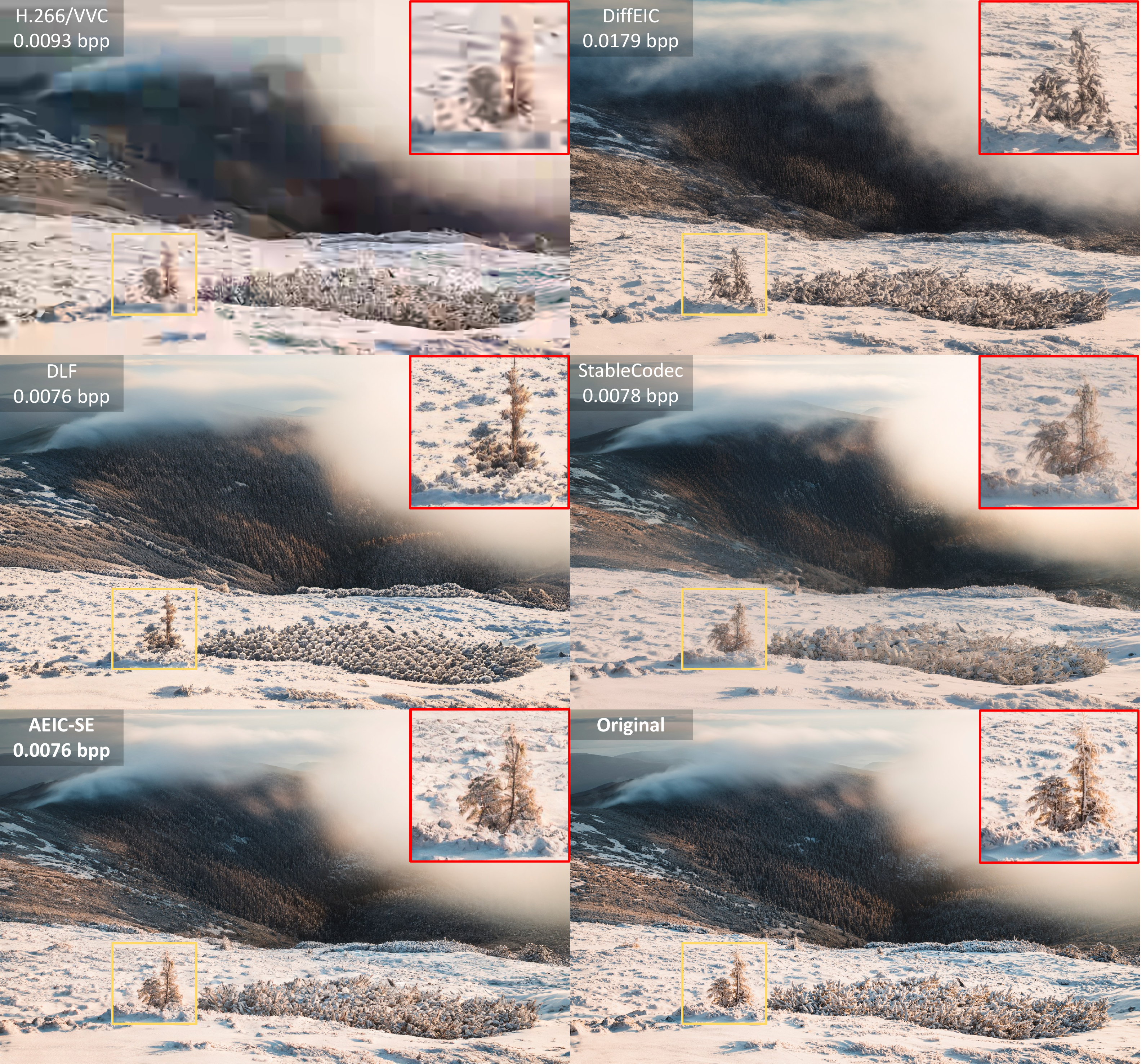}  
    \captionsetup{skip=2pt}
    \caption{Visual comparison (2K resolution) on the CLIC 2020 test set.}
    \label{visual5}
\end{figure*}

\begin{figure*}[!h]
    \centering
    \includegraphics[width=\textwidth]{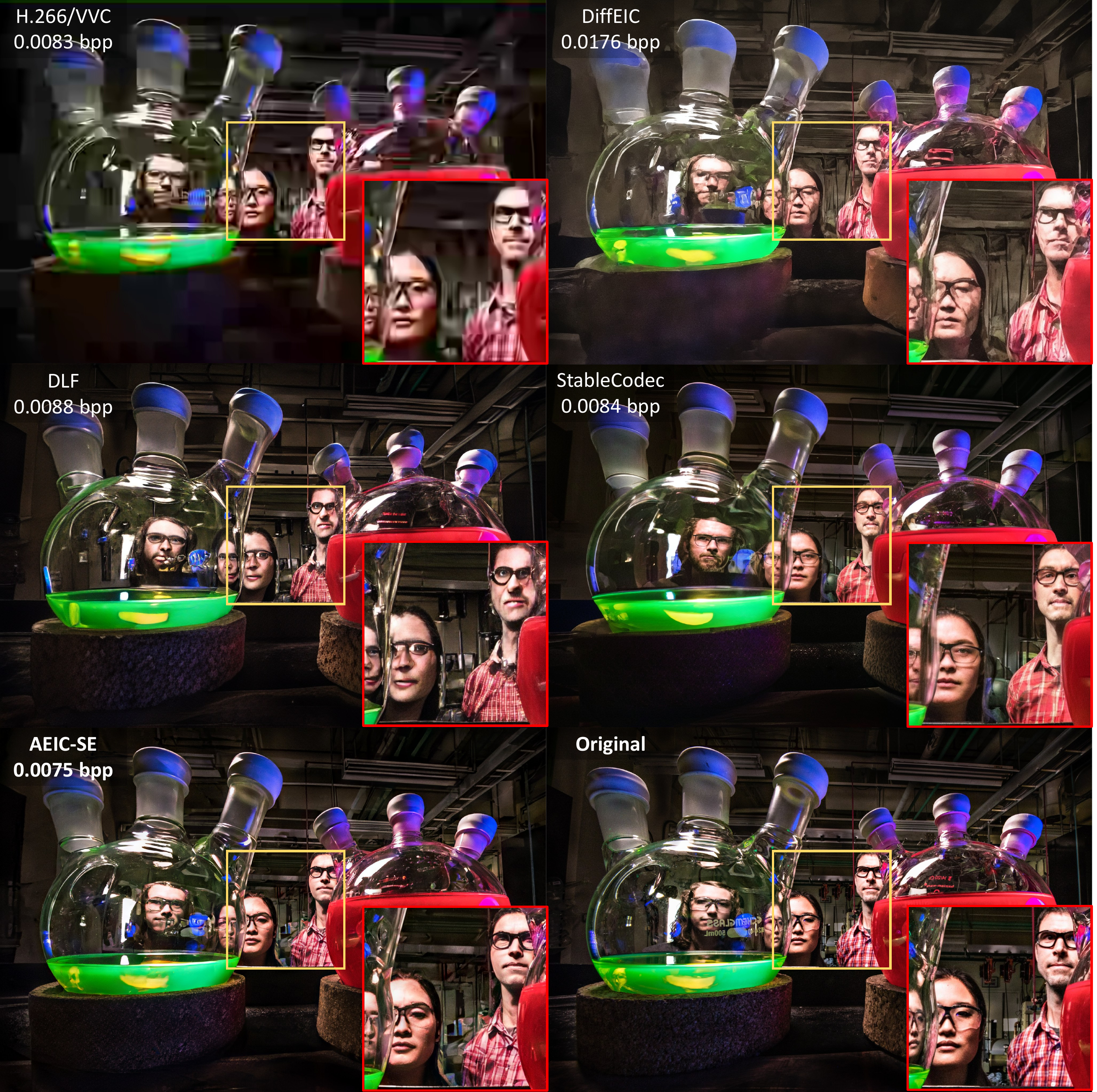}  
    \captionsetup{skip=2pt}
    \caption{Visual comparison (2K resolution) on the CLIC 2020 test set.}
    \label{visual6}
\end{figure*}

\begin{figure*}[!h]
    \centering
    \includegraphics[width=\textwidth]{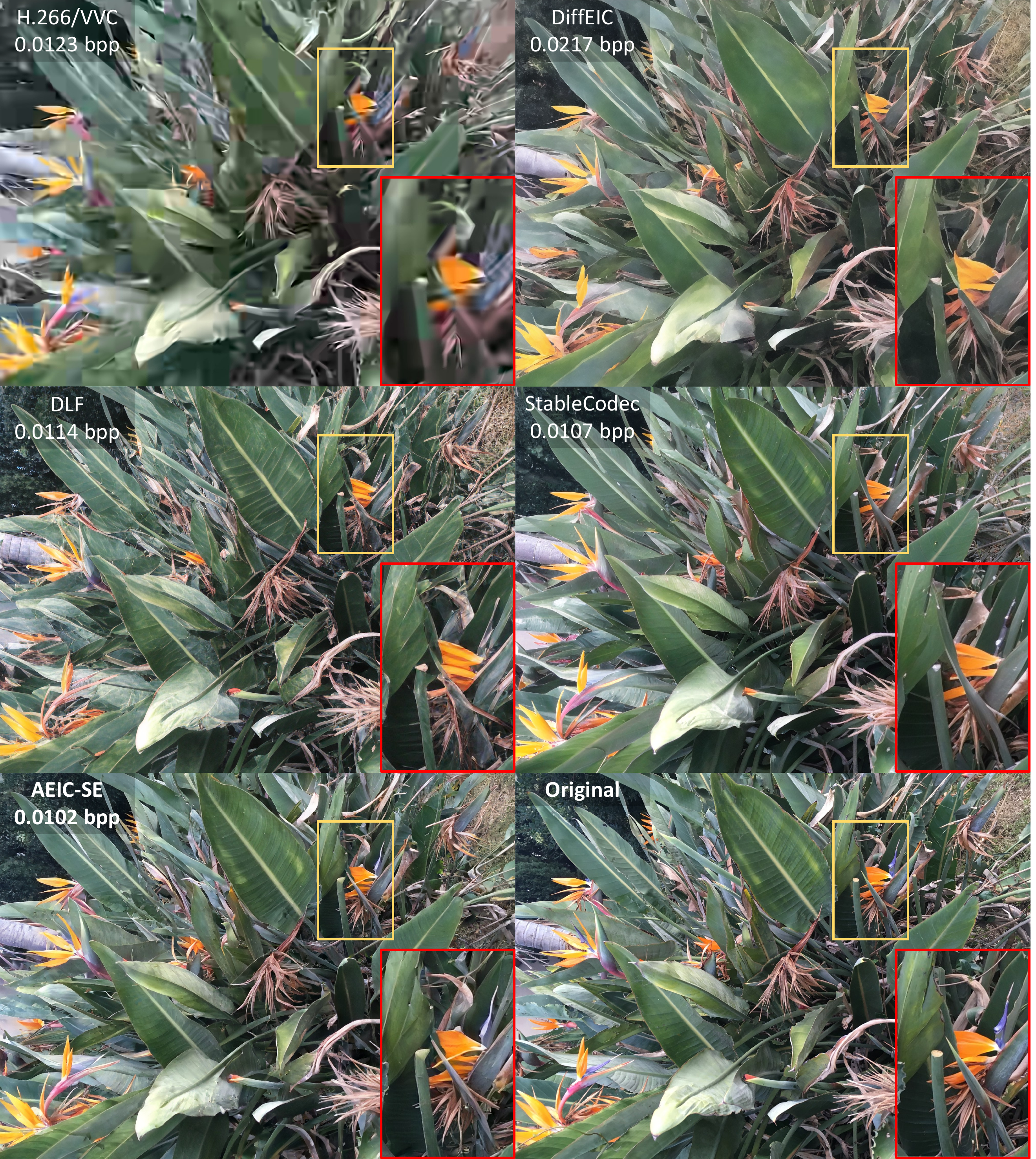}  
    \captionsetup{skip=2pt}
    \caption{Visual comparison (2K resolution) on the CLIC 2020 test set.}
    \label{visual7}
\end{figure*}


\end{document}